\documentclass[times,twocolumn,final,authoryear]{elsarticle}

\usepackage{ycviu}

\usepackage{hyperref}
\usepackage{framed,multirow}

\usepackage{amssymb}
\usepackage{latexsym}

\usepackage{url}
\usepackage{xcolor}
\definecolor{newcolor}{rgb}{.8,.349,.1}

\usepackage{cite}
\usepackage{amsmath}
\usepackage{amsfonts}
\usepackage{hhline}
\usepackage{makecell}
\usepackage{array}
\usepackage{tikz}
\usetikzlibrary{shapes.geometric, arrows, graphs, positioning, quotes, fit, decorations.pathreplacing}
\usepackage{rotating} 
\usepackage{subcaption}
\usepackage{caption}
\graphicspath{{pics/}}
\definecolor{darkcyan}{RGB}{15,144,144}
\definecolor{darkorange}{RGB}{202,112,5}
\definecolor{darkpurple}{RGB}{117,9,111}
\definecolor{lightred}{RGB}{230,29,72}
\definecolor{middlegreen}{RGB}{64,175,54}
\definecolor{darkgreen}{rgb}{0,0.6,0}
\definecolor{gray}{rgb}{0.5,0.5,0.5}
\definecolor{mauve}{rgb}{0.58,0,0.82}

\newcommand{\etal}{\emph{et al}.\@}

\journal{Computer Vision and Image Understanding}

\begin{document}

\ifpreprint
  \setcounter{page}{1}
\else
  \setcounter{page}{1}
\fi

\begin{frontmatter}

\title{Human-object interaction prediction in videos through gaze following}

\author[1]{Zhifan \snm{Ni}\corref{cor1}}
\ead{zhifan.ni@tum.de}
\author[2]{Esteve \snm{Valls Mascar\'o}}
\ead{esteve.valls.mascaro@tuwien.ac.at}
\author[3]{Hyemin \snm{Ahn}}
\ead{hyemin.ahn@unist.ac.kr}
\author[2,4]{Dongheui \snm{Lee}}
\ead{dongheui.lee@tuwien.ac.at}
\cortext[cor1]{Corresponding author}

\address[1]{Technical University of Munich (TUM), Arcisstr. 21, Munich 80333, Germany}
\address[2]{Technische Universität Wien (TU Wien), Karlsplatz 13, Vienna 1040, Austria}
\address[3]{Ulsan National Institute of Science and Technology (UNIST), UNIST-gil 50, Ulsan 44919, Republic of Korea}
\address[4]{German Aerospace Center (DLR), Muenchener Str. 20, Wessling 82234, Germany}

\begin{abstract}
    Understanding the human-object interactions (HOIs) from a video is essential to fully comprehend a visual scene. This line of research has been addressed by detecting HOIs from images and lately from videos. However, the video-based HOI anticipation task in the third-person view remains understudied. In this paper, we design a framework to detect current HOIs and anticipate future HOIs in videos. We propose to leverage human gaze information since people often fixate on an object before interacting with it. These gaze features together with the scene contexts and the visual appearances of human-object pairs are fused through a spatio-temporal transformer. To evaluate the model in the HOI anticipation task in a multi-person scenario, we propose a set of person-wise multi-label metrics. Our model is trained and validated on the VidHOI dataset, which contains videos capturing daily life and is currently the largest video HOI dataset. Experimental results in the HOI detection task show that our approach improves the baseline by a great margin of 36.3\% relatively. Moreover, we conduct an extensive ablation study to demonstrate the effectiveness of our modifications and extensions to the spatio-temporal transformer. Our code is publicly available on \url{https://github.com/nizhf/hoi-prediction-gaze-transformer}.
\end{abstract}

\end{frontmatter}



\section{Introduction}

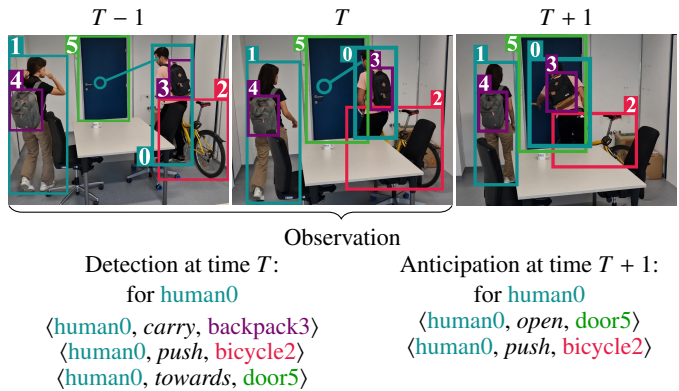
\begin{figure}
    \centering
    \begin{tikzpicture}[font=\small, >={latex}]
        \node[inner sep=0pt] (f__1) {\includegraphics[width=0.158\textwidth]{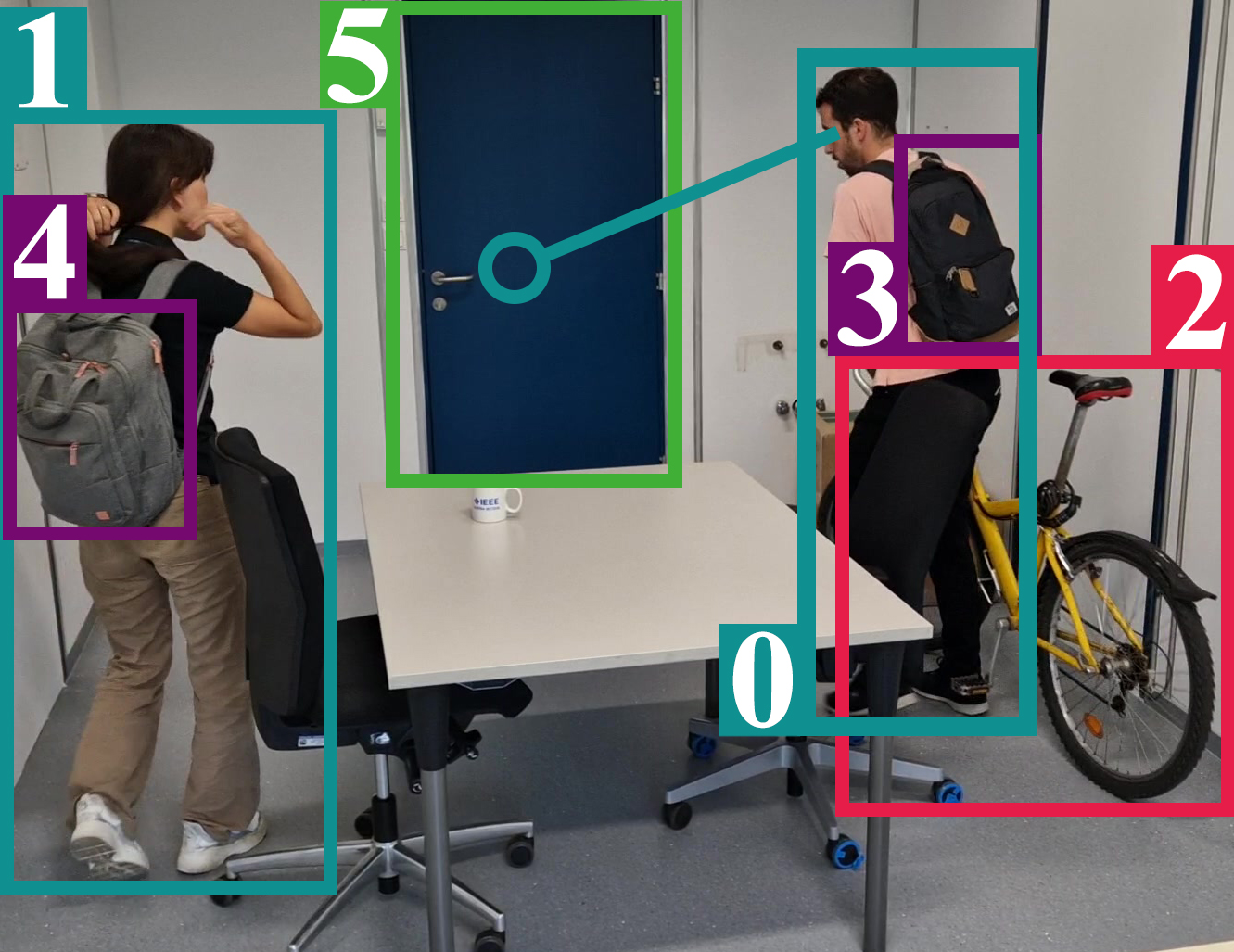}};
        \node[inner sep=0pt, right=1pt of f__1] (f_0) {\includegraphics[width=0.158\textwidth]{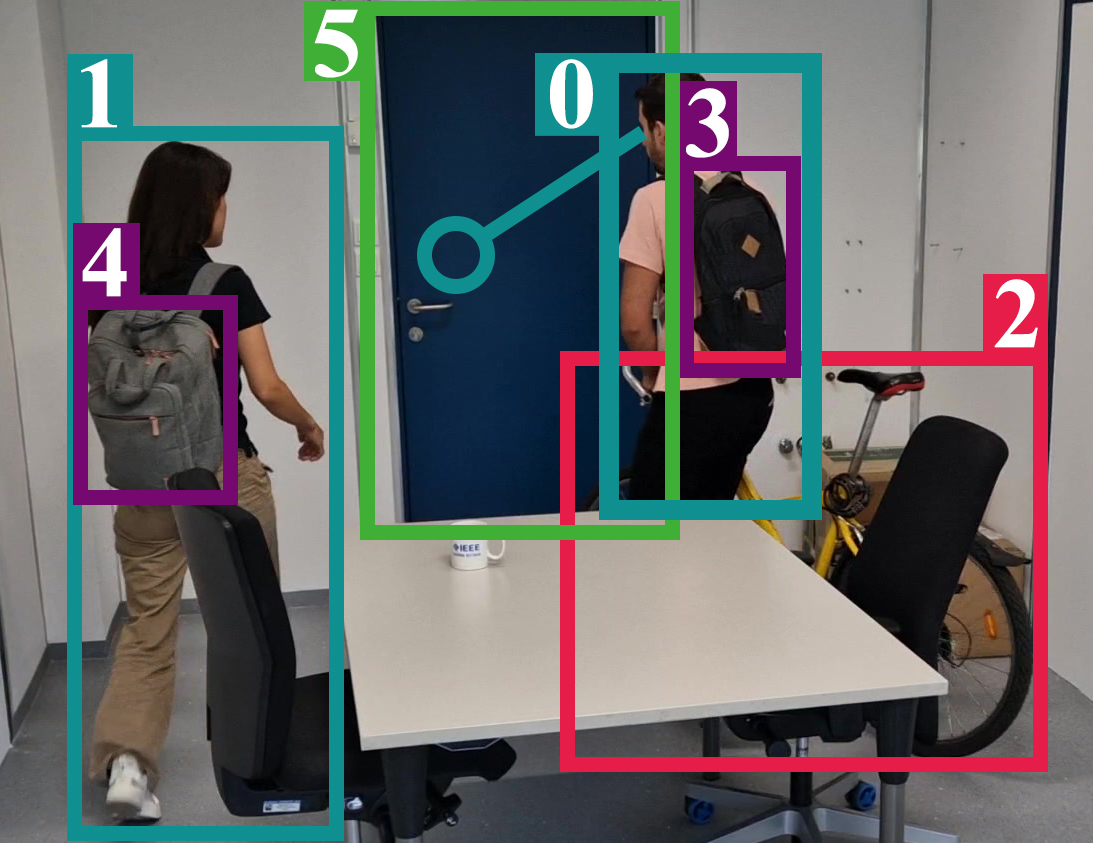}};
        \node[inner sep=0pt, right=1pt of f_0] (f_1) {\includegraphics[width=0.158\textwidth]{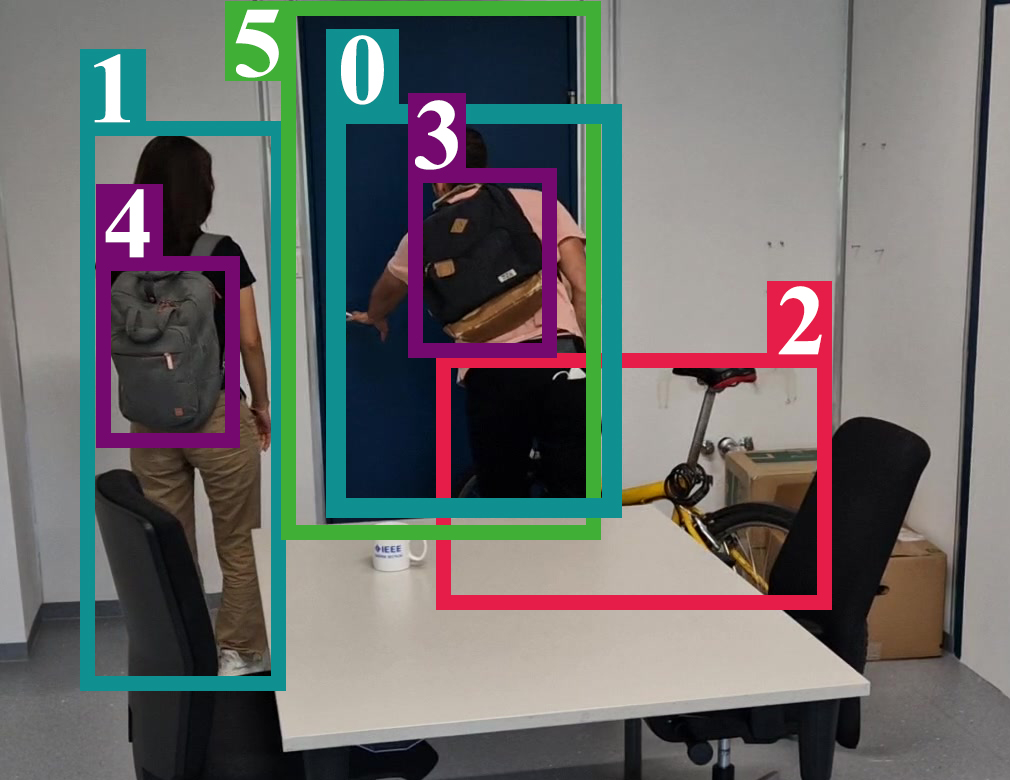}};

        \node[draw=none, rectangle, align=center, above=0pt of f__1] (t__1) {$T-1$};
        \node[draw=none, rectangle, align=center, above=0pt of f_0] (t_0) {$T$};
        \node[draw=none, rectangle, align=center, above=0pt of f_1] (t_1) {$T+1$};

        \draw [decorate, decoration={brace,amplitude=5pt,mirror,aspect=0.75}] ([xshift=-41pt]f__1.south) -- ([xshift=41pt]f_0.south) node[midway, below=1pt of f__1] {};

        \node[draw=none, rectangle, align=center, below=5pt of f_0] (tt) {Observation};

        \node[draw=none, rectangle, align=center, below=-3pt of tt, xshift=-60pt] (det) {Detection at time $T$:\\for \textcolor{darkcyan}{human0} };
        \node[draw=none, inner sep=0pt, rectangle, align=center, below=2pt of det] (d1) {$\langle$\textcolor{darkcyan}{human0}, \emph{carry}, \textcolor{darkpurple}{backpack3}$\rangle$};
        \node[draw=none, inner sep=0pt, rectangle, align=center, below=1pt of d1] (d2) {$\langle$\textcolor{darkcyan}{human0}, \emph{push}, \textcolor{lightred}{bicycle2}$\rangle$};
        \node[draw=none, inner sep=0pt, rectangle, align=center, below=1pt of d2] (d3) {$\langle$\textcolor{darkcyan}{human0}, \emph{towards}, \textcolor{darkgreen}{door5}$\rangle$};

        \node[draw=none, rectangle, align=center, below=-3pt of tt, xshift=70pt] (ant) {Anticipation at time $T+1$:\\for \textcolor{darkcyan}{human0}};
        \node[draw=none, inner sep=0pt, rectangle, align=center, below=0pt of ant] (a1) {$\langle$\textcolor{darkcyan}{human0}, \emph{open}, \textcolor{darkgreen}{door5}$\rangle$};

        \node[draw=none, inner sep=0pt, rectangle, align=center, below=1pt of a1] (a2) {$\langle$\textcolor{darkcyan}{human0}, \emph{push}, \textcolor{lightred}{bicycle2}$\rangle$};
        
    \end{tikzpicture}
    \caption{An example of HOI detection and anticipation tasks with gaze-following method from a video. By observing a past sequence of RGB frames, the model should detect current HOIs or forecast possible HOIs after one second. The gaze cues provide information about human attention and are useful to determine which object is more likely to be interacted with in a complex scene. }
    \label{fig:intro}
\end{figure}

Detecting human-object interactions (HOIs) is a fundamental step toward high-level comprehension of scenes. Compared to instance-level visual recognition tasks such as object detection~\citep{detection:faster_rcnn, detection:detr, detection:yolov5} and action recognition~\citep{action:two_stream, action:kinetics}, HOI detection can provide more contextual and fine-grained cues for scene understanding. However, real-world applications, such as robotics, autonomous driving, and surveillance system, usually need to reason about a scene and generate a plausible HOI anticipation for the near future. For instance, as shown in Fig.~\ref{fig:intro}, the person on the right is \emph{pushing} a bicycle and \emph{walking towards} a door. Based on this observation, if an intelligent system could anticipate that the human will open the door, it could assist that person to perform this interaction beforehand. Then the human could leave the room without interruption. Thus, a framework that can forecast future HOIs from a video is essential. 

However, HOI detection and anticipation are still challenging as multiple humans and objects may appear in a scene and a human may have multiple interactions with multiple objects. In addition, the dependencies between frames are crucial to understand the temporal evolution of human interactions. Due to these difficulties, most existing approaches are only designed for HOI detection in static images. Conventional methods~\citep{hoi_i2:visual_semantic, hoi_i2:language_prior, hoi_i2:learning_hico, hoi_i2:ican, hoi_i2:detecting_recognizing, hoi_i2:neural_motifs, hoi_i2:reidn, hoi_i2:interact_intend, hoi_i2:gpsnet} often contain two stages. First, an object detector is applied to locate humans and objects. Second, a multi-stream classifier predicts the interactions for each human-object pair. To increase the model efficiency, several one-stage or end-to-end methods~\citep{hoi_i1:learning_interaction_point, hoi_i1:ppdm, hoi_i1:hotr, hoi_i1:qpic} are proposed to generate object detection and interaction classes in parallel.

While the image-based HOI detectors show great performance on image datasets, they may perform poorly on video datasets because they cannot exploit the temporal cues required to distinguish between some continuous interactions, such as \emph{open} or \emph{close} a door~\citep{hoi_v_set:vlog}. Hence, a few works~\citep{hoi_v2:learning_gpn, hoi_v_set:VidHOI, hoi_v2:sttran, hoi_v2:detecting_hort, hoi_v2:st_gpn, hoi_v2:tubelet_tokens} are proposed to leverage the temporal dependencies between frames and demonstrate superior performance to the image-based methods. However, these approaches do not consider the human gaze as an additional feature while it often provides valuable information about human intentions~\citep{eye:eye_hand_coordination, eye:what_ways, eye:visual_memory, eye:landscape, eye:planning_precise}.

To enable an intelligent system to collaborate with humans more effectively, only recognizing the current HOIs is not sufficient. The ability to anticipate subsequent HOIs is beneficial for task planning and danger avoidance. Nevertheless, there are very few studies addressing the HOI anticipation task from the third-person view~\citep{hoi_v_anti:srnn, hoi_v_anti:red, hoi_v_anti:structured_lstm, hoi_v_anti:lighten}. However, these works are conducted on small-scale datasets and cannot be generalized to real-world applications. 

Thus, we propose a multimodal framework that leverages visual appearance features, semantic contexts, and human gaze cues to tackle HOI detection and anticipation tasks in videos. To our best knowledge, our work is the first one attempting to utilize gaze features in video-based HOI anticipation, and the first to anticipate HOIs in multi-person scenarios. Our framework works in two-stage as follows: in the first stage, an object module detects and tracks humans and objects across the video, and a gaze module leverages human head features to identify where the human is looking at every instant. In the second stage, a spatio-temporal transformer aggregates all extracted features from a sliding window of frames to infer the current or future HOIs. Our spatio-temporal transformer is inspired by the STTran model~\citep{hoi_v2:sttran}. However, we observe several limitations in STTran architecture that diminish the performance. First, we notice that using the spatial encoder to implicitly extract intra-frame contexts yields a very small benefit. Since the global scene context is useful for vision-related tasks~\citep{hoi_i2:deep_contextual, detection:global_context_aware, detection:reasonable_global_context}, we extend the spatial encoder to explicitly generate a global feature vector for each frame. Inspired by Vision Transformer (ViT)~\citep{transformer:vit}, we prepend a learnable class token to the spatial encoder input, which captures the global relationship among all human-object pairs at a particular moment. Moreover, we observe that the temporal encoder in STTran infers temporal relations of all human-object pairs from a sliding window of frames. Instead, we propose an instance-level temporal encoder, which independently processes each unique human-object pair. This allows our model to focus on the individual evolution of each human-object representation in time. Finally, we apply a cross-attention layer to fuse the extracted global features and the gaze information with the instance-level human-object representations. Therefore, our architecture proposes a big extension to STTran and clearly boosts its performance in both HOI detection and anticipation tasks.

Our model is trained and validated on VidHOI dataset~\citep{hoi_v_set:VidHOI}, which is composed of daily-life videos and is currently the largest video HOI dataset. We design a training strategy to address the dataset imbalance issue. Moreover, inspired by the metrics for egocentric action anticipation tasks, we propose a set of person-wise metrics to assess the model in the HOI anticipation task on multi-person videos. These metrics compute the multi-label recall, precision, accuracy, and F1-score~\citep{metric:multi_label} separately for each human using the top-$k$ predictions. We also conduct an extensive ablation study to confirm the effectiveness of our modified and added components. 

The main contributions of our work are summarized as:
\begin{enumerate}
    \item A deep multimodal spatio-temporal transformer network is designed for anticipating HOIs in multi-person scenes.
    \item The use of gaze-following methodology in the cross-attention mechanism is explored as an additional novel step towards HOI detection and anticipation in videos.   
    \item A person-wise multi-label criterion is proposed to evaluate the HOI anticipation model in third-person videos.
\end{enumerate}

\section{Related Works}

\subsection{Gaze in HOI Detection}
A Human's gaze direction can indicate where the human is paying attention to. Cognitive studies~\citep{eye:what_ways, eye:visual_memory} show that human eyes often fixate on the object when performing manual actions with it. Moreover, humans sometimes move their gaze to the next object before finishing the current interaction. \citet{eye:landscape} further suggest that humans may scan over all task-relevant objects when planning a complex movement. \citet{eye:planning_precise} then discover that the gaze point on an object is dependent on the interaction type. The above-mentioned works demonstrate that gaze cues can provide useful information for detecting and anticipation HOIs. 

However, the use of gaze features in HOI detection is not much investigated. For image-based HOI detection, \citet{hoi_i2:interact_intend} propose a human intention-driven HOI detection framework, which utilizes human pose and gaze to assist HOI detection. Their ablation study shows that utilizing human gaze regions can improve the model performance. Nevertheless, to the best of our knowledge, there is no work leveraging the human gaze in video-based HOI tasks. To bridge this gap, our framework explores the effectiveness of gaze information in HOI detection and HOI anticipation tasks. 

\subsection{Video-based HOI Detection}
To properly detect interactions between a human and an object from a video, understanding the evolution of the pair relationship over time is essential. For instance, \citet{hoi_v_anti:srnn} represent human-object relations as a spatio-temporal graph and adopts a Structural Recurrent Neural Network (S-RNN) to infer the interaction types. \citet{hoi_v_anti:structured_lstm} refine the S-RNN by additionally considering object-object relations. \citet{hoi_v_anti:lighten} further improve the model performance by applying learned visual features as the graph nodes. Instead of RNNs, \citet{hoi_v2:learning_gpn} propose a Graph Parsing Network (GPN) to parse the spatio-temporal graphs of human-object interactions. Then, \citet{hoi_v2:st_gpn} design a two-stream GPN that also incorporates the semantic features. In contrast to the graph-based methods, \citet{hoi_v2:sthoid} propose an instance-based architecture to separately reason each human-object pair instance. This model leverages human skeletons as an additional cue for HOIs. ST-HOI~\citep{hoi_v_set:VidHOI} also utilizes human pose features to detect HOIs. In addition, ST-HOI applies a 3D backbone to extract correctly-localized instance features from a video. Moreover, the large-scale VidHOI dataset is proposed to enable the development of large-size models. Recently, motivated by the great success of the transformer model, different instance-based spatio-temporal transformers~\citep{hoi_v2:detecting_hort, hoi_v2:sttran, hoi_v2:tubelet_tokens} are designed and are reviewed in the next section.

\subsection{Transformer in HOI Detection}
The transformer~\citep{transformer:vanilla} is designed for natural language processing (NLP) tasks. The key component in transformer is the attention mechanism, which copes with the gradient vanishing problem of recurrent neural networks (RNNs) in long data sequences. In many NLP tasks, transformer models outperform RNN-based models by a great margin. 

Recent advances in transformer in computer vision tasks have motivated researchers to apply it also in the HOI detection task. Several approaches~\citep{hoi_i1:hotr, hoi_i1:qpic, hoi_i1:structure_aware_trans, hoi_i1:consistency_learning, hoi_i1:doq} attempt to extend the Detection Transformer (DETR)~\citep{detection:detr} from object detection to HOI detection in static images. These approaches first use a convolutional neural network (CNN) to extract visual features from the input image. Then, a transformer network aggregates image-wide contextual features and returns the human bounding box, object bounding box, object class, and interaction class in parallel. These models achieve state-of-the-art performance in the image-based HOI detection task. However, they may perform poorly when detecting HOIs in a video as they cannot understand the temporal contexts between frames. 

Recently, researchers~\citep{hoi_v2:detecting_hort, hoi_v2:sttran, hoi_v2:tubelet_tokens} propose to detect HOIs from videos using spatio-temporal transformers. \citet{hoi_v2:detecting_hort} design the Human-Object Relationship Transformer (HORT), which leverages both visual appearance and human pose features to facilitate HOI detection. These features are fused by a transformer with densely-connected parallel spatial and temporal encoders. In contrast, Spatial-Temporal Transformer (STTran)~\citep{hoi_v2:sttran} consists of a sequential architecture of spatial and temporal transformer encoders. The visual appearance feature of each human-object instance is concatenated with the spatial relation feature and the semantic feature. Most recently, inspired by ViT~\citep{transformer:vit}, \citet{hoi_v2:tubelet_tokens} extract patch tokens from frames by a spatial encoder and link them to tubelet tokens across time. A transformer decoder similar to DETR~\citep{detection:detr} reasons HOIs from the tubelet tokens by using learned positional encodings.

Nevertheless, the above-mentioned spatio-temporal models do not consider gaze cues, which could provide useful information for HOI detection and anticipation. Thus, we introduce gaze features as an additional modality to a spatial-temporal transformer model. We choose STTran~\citep{hoi_v2:sttran} as our base model since it achieves remarkable performance on the Action Genome~\citep{hoi_v_set:action_genome} dataset and can be easily extended with more features.

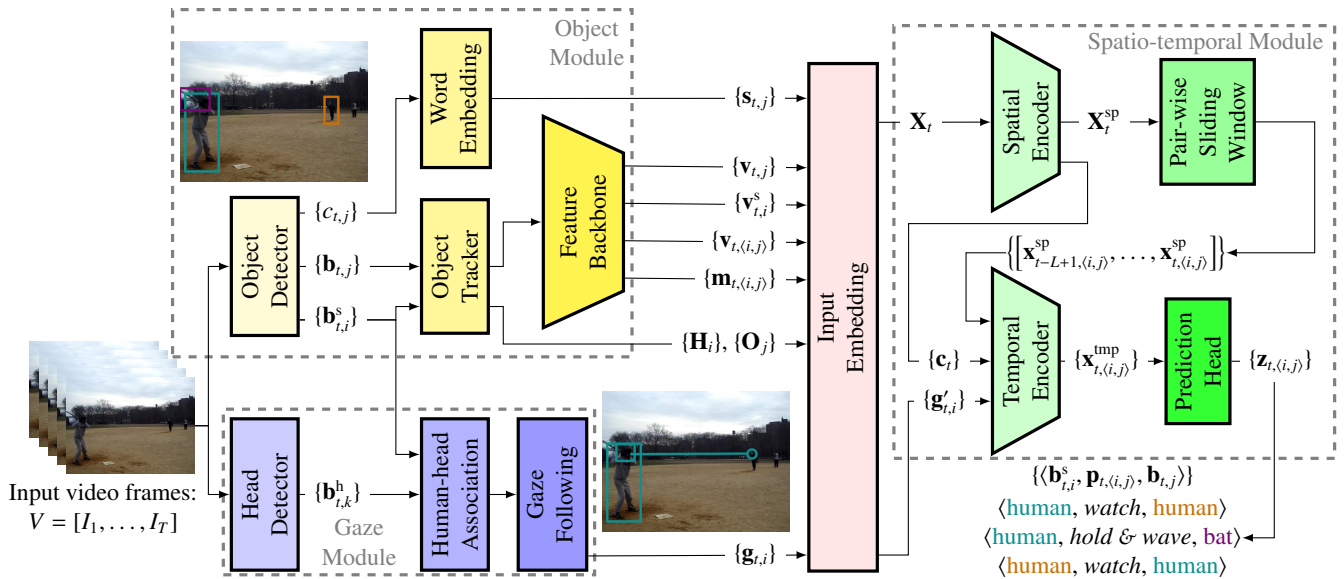
\begin{figure*}
    \centering
    \begin{tikzpicture}[font=\small, >={latex}]
        \node[inner sep=0pt] (f5) {\includegraphics[width=40pt]{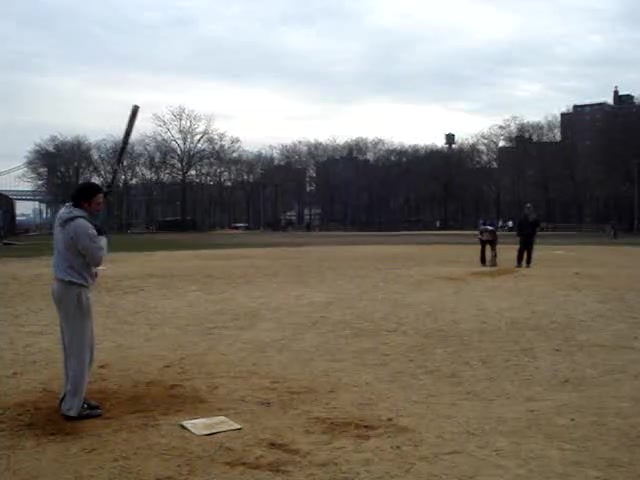}};
        \node[inner sep=0pt, below right=-33pt and -45pt of f5] (f4) {\includegraphics[width=48pt]{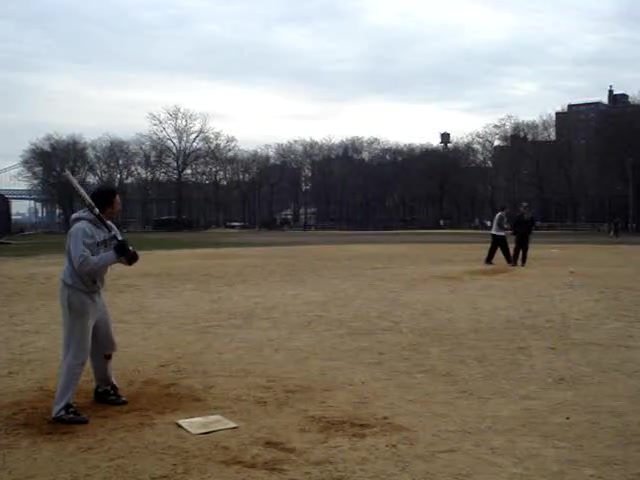}};
        \node[inner sep=0pt, below right=-33pt and -45pt of f4] (f3) {\includegraphics[width=48pt]{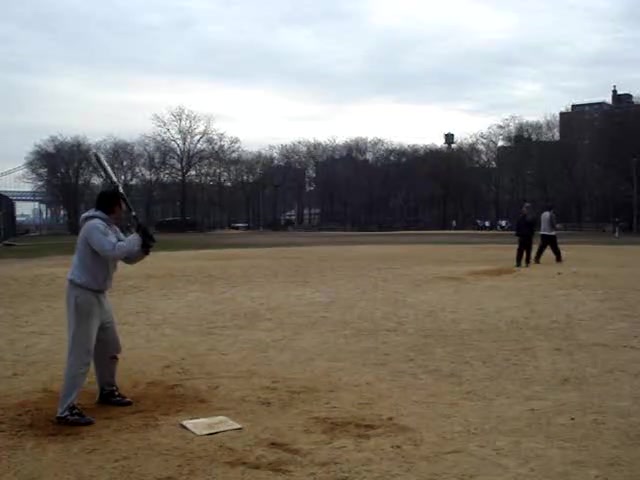}};
        \node[inner sep=0pt, below right=-33pt and -45pt of f3] (f2) {\includegraphics[width=48pt]{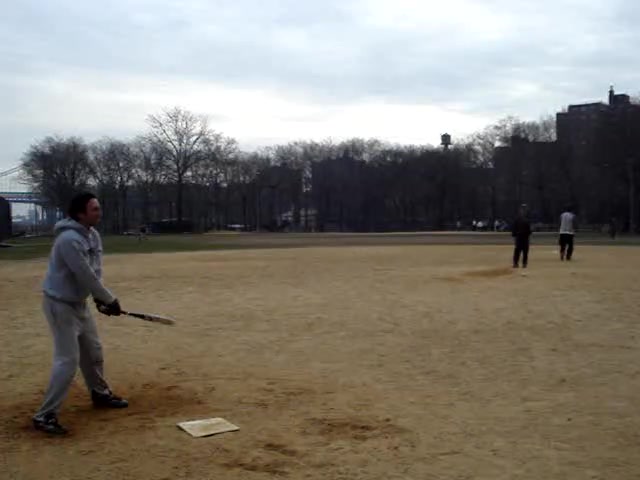}};
        \node[inner sep=0pt, below right=-33pt and -45pt of f2] (f1) {\includegraphics[width=48pt]{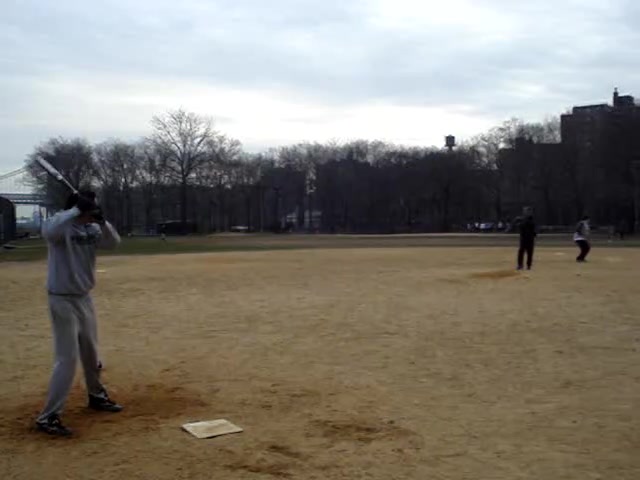}};
        \node[inner sep=0pt, below right=-33pt and -45pt of f1] (f0) {\includegraphics[width=48pt]{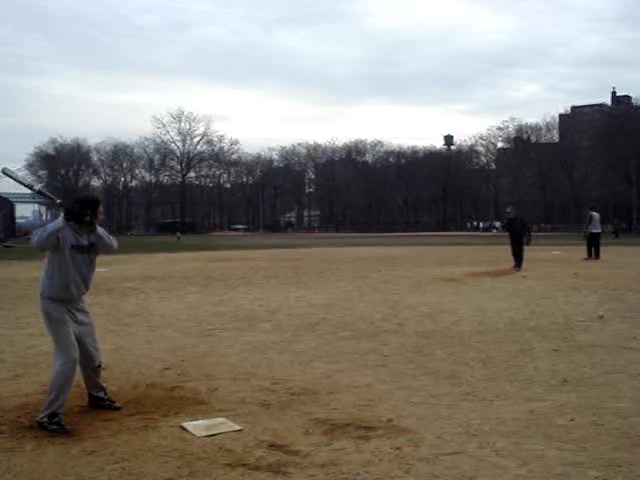}};
        \node[draw=none, rectangle, align=center, below=10pt of f3] (in) {Input video frames:\\$V=[I_1, \dots, I_T]$};
  
        \node[draw, rectangle, align=center, fill=yellow!20, very thick, minimum height=45pt, minimum width=25pt, above right=15pt and 13pt of f0] (obj) {\rotatebox{90}{\parbox[c]{45pt}{\centering Object\\Detector}}};
        \node[draw, rectangle, align=center, fill=yellow!50, very thick, minimum height=50pt, minimum width=25pt, right=45pt of obj] (track) {\rotatebox{90}{\parbox[c]{35pt}{\centering Object\\Tracker}}};
        \node[draw, rectangle, align=center, fill=yellow!50, very thick, minimum height=50pt, minimum width=25pt, above right=10pt and 45pt of obj] (word) {\rotatebox{90}{\parbox[c]{45pt}{\centering Word\\Embedding}}};
        \node[draw, trapezium, align=center, fill=yellow!80, very thick, anchor=north, rotate=-90, minimum height=30pt, trapezium stretches body, right=35pt of track, xshift=-50pt] (back) {\rotatebox{180}{\parbox[c]{45pt}{\centering Feature\\Backbone}}};
        \node[inner sep=0pt, above right=5pt and -45pt of obj] (fobj) {\includegraphics[width=70pt]{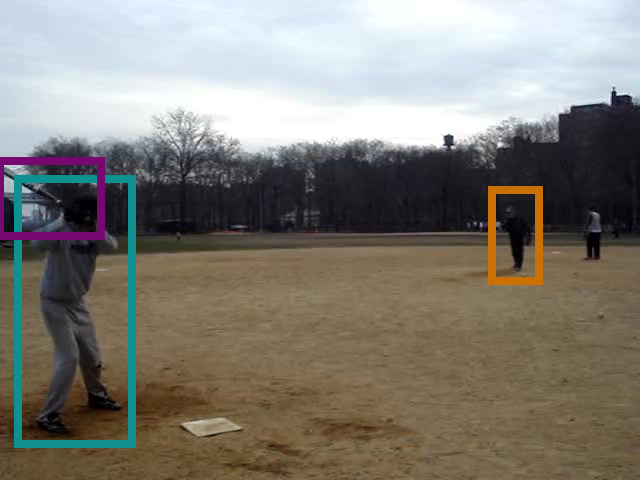}};
        \node[draw, color=gray, rectangle, align=center, fill=none, dashed, very thick, minimum height=130pt, minimum width=172pt, above right=7pt and -9pt of f0] (obj_m) {};
        \node[draw=none, rectangle, align=center, fill=none, above right=-25pt and -48pt of obj_m] {\parbox[c]{6em}{\centering\textcolor{gray}{Object\\Module}}};
        \node[draw, rectangle, align=center, fill=blue!20, very thick, minimum height=50pt, minimum width=25pt, below=30pt of obj] (head) {\rotatebox{90}{\parbox[c]{50pt}{\centering Head\\Detector}}};
        \node[draw, rectangle, align=center, fill=blue!30, very thick, minimum height=50pt, minimum width=25pt, right=45pt of head] (asso) {\rotatebox{90}{\parbox[c]{50pt}{\centering Human-head\\Association}}};
        \node[draw, rectangle, align=center, fill=blue!40, very thick, minimum height=50pt, minimum width=25pt, right=10pt of asso] (gaze) {\rotatebox{90}{\parbox[c]{50pt}{\centering Gaze\\Following}}};
        \node[inner sep=0pt, above right=-43pt and 5pt of gaze] (fgaze) {\includegraphics[width=70pt]{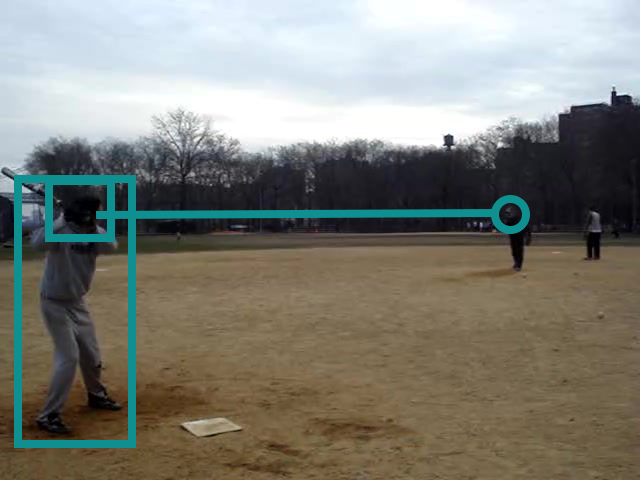}};
        \node[draw, color=gray, rectangle, align=center, fill=none, dashed, very thick, minimum height=62pt, minimum width=139pt, below right=-25pt and 10pt of f0] (gaze_m) {};
        \node[draw=none, rectangle, align=center, fill=none, below left=-25pt and -82pt of gaze_m] {\parbox[c]{6em}{\centering\textcolor{gray}{Gaze\\Module}}};
        \node[draw, rectangle, align=center, fill=pink!40, very thick, minimum height=190pt, minimum width=25pt, below right=-103pt and 190pt of obj] (emb) {\rotatebox{90}{\parbox[c]{50pt}{\centering Input\\Embedding}}};
        \node[draw, trapezium, align=center, fill=green!20, very thick, anchor=north, rotate=-90, minimum height=25pt, trapezium stretches body, right=55pt of emb, xshift=-100pt] (spa) {\rotatebox{180}{\parbox[c]{32pt}{\centering Spatial\\Encoder}}};
        \node[draw, rectangle, align=center, fill=green!40, very thick, minimum height=40pt, minimum width=20pt, right=50pt of spa, yshift=27pt] (concat) {\rotatebox{90}{\parbox[c]{40pt}{\centering Pair-wise\\Sliding\\Window}}};
        \node[draw, trapezium, align=center, fill=green!20, very thick, anchor=north, rotate=-90, minimum height=25pt, trapezium stretches body, right=55pt of emb, xshift=-12pt] (tmp) {\rotatebox{180}{\parbox[c]{35pt}{\centering Temporal\\Encoder}}};
        \node[draw, rectangle, align=center, fill=green!80, very thick, minimum height=40pt, minimum width=20pt, below right=-37pt and 65pt of tmp] (pred) {\rotatebox{90}{\parbox[c]{40pt}{\centering Prediction\\Head}}};
        \node[draw, color=gray, rectangle, align=center, fill=none, dashed, very thick, minimum height=162pt, minimum width=165pt, above right=-149pt and 5pt of emb] (st_m) {};
        \node[draw=none, rectangle, align=center, fill=none, above left=-15pt and -165pt of st_m] {\centering\textcolor{gray}{Spatio-temporal Module}};
        \node[draw=none, rectangle, align=center, inner xsep=0, below left=15pt and -25pt of concat] (s1) {$\left\{\left[\mathbf{x}_{t-L+1,\langle i,j \rangle}^\text{sp}, \dots, \mathbf{x}_{t,\langle i,j \rangle}^\text{sp}\right]\right\}$};
        \node[draw=none, inner sep=0pt, rectangle, align=center, below right=15pt and -75pt of pred] (outg) {$\{\langle \mathbf{b}_{t,i}^\text{s}, \mathbf{p}_{t,\langle i,j \rangle}, \mathbf{b}_{t,j} \rangle\}$};
        \node[draw=none, inner sep=0pt, rectangle, align=center, below=2pt of outg] (out1) {
        $\langle$\textcolor{darkcyan}{human}, \emph{watch}, \textcolor{darkorange}{human}$\rangle$ \\
        $\langle$\textcolor{darkcyan}{human}, \emph{hold \& wave}, \textcolor{darkpurple}{bat}$\rangle$ \\
        $\langle$\textcolor{darkorange}{human}, \emph{watch}, \textcolor{darkcyan}{human}$\rangle$};
  
        \draw[->] (f0.east) --++ (5pt, 0) |- (obj.west);
        \draw[->] ([yshift=20pt]obj.east) --++ (36pt, 0) |- (word.west);
        \draw[->] ([yshift=0pt]obj.east) -> (track.west);
        \draw[->] ([yshift=-20pt]obj.east) --++ (36pt, 0) |- ([yshift=-15pt]track.west);
        \draw[->] ([yshift=-20pt]obj.east) --++ (36pt, 0) |- ([yshift=15pt]asso.west);
        \draw[->] ([yshift=0]word.east) -> ([yshift=82pt]emb.west);
        \draw[->] (track.east) --++ (5pt, 0) |- (back.south);
        \draw[->] ([yshift=-15pt]track.east) --++ (5pt, 0) |- ([yshift=-10pt]emb.west);
        \draw[->] ([yshift=-21pt]back.north) -> ([yshift=14pt]emb.west);
        \draw[->] ([yshift=-7pt]back.north) -> ([yshift=28pt]emb.west);
        \draw[->] ([yshift=7pt]back.north) -> ([yshift=42pt]emb.west);
        \draw[->] ([yshift=21pt]back.north) -> ([yshift=56pt]emb.west);
  
        \draw[->] (f0.east) --++ (5pt, 0) |- (head.west);
        \draw[->] (head.east) -> (asso.west);
        \draw[->] (asso.east) -> (gaze.west);
        \draw[->] ([yshift=-23pt]gaze.east) -> ([yshift=-89pt]emb.west);
  
        \draw[->] ([yshift=73pt]emb.east) -> (spa.south);
        \draw[->] ([yshift=-89pt]emb.east) --++ (11pt, 0) |- ([yshift=-15pt]tmp.south);
  
        \draw[->] (spa.north) -> (concat.west);
        \draw[->] ([yshift=-15pt]spa.north) --++ (10pt, 0) --++ (0, -23pt) --++ (-67pt, 0) |- (tmp.south);
        \draw[->] (concat.east) --++ (22pt, 0) |- (s1.east);
        \draw[->] (s1.west) --++ (-15pt, 0) |- ([yshift=15pt]tmp.south);
        \draw[->] (tmp.north) -> (pred.west);
        \draw[->] (pred.east) --++ (16pt, 0) |- (out1.east);
  
        \node[draw=none, rectangle, align=center, fill=white, right=2pt of obj, yshift=20pt] (cls) {$\{c_{t,j}\}$};
        \node[draw=none, rectangle, align=center, fill=white, right=2pt of obj, yshift=0pt] (obj_d) {$\{\mathbf{b}_{t,j}\}$};
        \node[draw=none, rectangle, align=center, fill=white, right=2pt of obj, yshift=-20pt] (subj_d) {$\{\mathbf{b}_{t,i}^\text{s}\}$};
        \node[draw=none, rectangle, align=center, fill=white, right=2pt of head, yshift=0pt] (head_d) {$\{\mathbf{b}_{t,k}^\text{h}\}$};
  
        \node[draw=none, rectangle, align=center, fill=white, left=8pt of emb, yshift=82pt] (v_s) {$\{\mathbf{s}_{t,j}\}$};
        \node[draw=none, rectangle, align=center, fill=white, left=8pt of emb, yshift=56pt] (v_o) {$\{\mathbf{v}_{t,j}\}$};
        \node[draw=none, rectangle, align=center, fill=white, left=8pt of emb, yshift=42pt] (v_h) {$\{\mathbf{v}_{t,i}^\text{s}\}$};
        \node[draw=none, rectangle, align=center, fill=white, left=8pt of emb, yshift=28pt] (v_r) {$\{\mathbf{v}_{t,\langle i,j \rangle}\}$};
        \node[draw=none, rectangle, align=center, fill=white, left=8pt of emb, yshift=14pt] (v_m) {$\{\mathbf{m}_{t,\langle i,j \rangle}\}$};
        \node[draw=none, rectangle, align=center, fill=white, left=8pt of emb, yshift=-89pt] (v_g) {$\{\mathbf{g}_{t,i}\}$};
        \node[draw=none, rectangle, align=center, fill=white, left=8pt of emb, yshift=-10pt] (H) {$\{\mathbf{H}_i\}$, $\{\mathbf{O}_j\}$};
  
        \node[draw=none, rectangle, align=center, fill=white, left=21pt of tmp, yshift=-28pt] (v_c) {$\{\mathbf{c}_{t}\}$};
        \node[draw=none, rectangle, align=center, fill=white, left=19pt of tmp, yshift=-44pt] (v_gp) {$\{\mathbf{g}^\prime_{t,i}\}$};
  
        \node[draw=none, rectangle, align=center, fill=white, right=8pt of emb, yshift=73pt] (X) {$\mathbf{X}_t$};
        \node[draw=none, rectangle, align=center, fill=white, left=11pt of concat] (Xs) {$\mathbf{X}_t^\text{sp}$};
        \node[draw=none, rectangle, align=center, fill=white, left=6pt of pred, yshift=0pt] (Xt1) {$\{\mathbf{x}_{t,\langle i,j \rangle}^\text{tmp}\}$};
        \node[draw=none, rectangle, align=center, fill=white, right=3pt of pred, yshift=0pt] (Z1) {$\{\mathbf{z}_{t,\langle i,j \rangle}\}$};
  
    \end{tikzpicture}
  
    \caption{Overview of our video-based HOI detection and anticipation framework. The framework consists of three modules. The object module detects bounding boxes of humans $\{\mathbf{b}_{t,i}^\text{s}\}$ and objects $\{\mathbf{b}_{t,j}\}$, and recognizes object classes $\{c_{t,j}\}$. An object tracker obtains human and object trajectories ($\{\mathbf{H}_i\}$ and $\{\mathbf{O}_j\}$) in the video. Then, the human visual features $\{\mathbf{v}_{t,i}^\text{s}\}$, object visual features $\{\mathbf{v}_{t,j}\}$, visual relation features $\{\mathbf{v}_{t,\langle i,j \rangle}\}$, and spatial relation features $\{\mathbf{m}_{t,\langle i,j \rangle}\}$ are extracted through a feature backbone. In addition, a word embedding model~\citep{semantic:glove} is applied to generate semantic features $\{\mathbf{s}_{t,j}\}$ of the object class. Meanwhile, the gaze module detects heads $\{\mathbf{b}_{t,k}^\text{h}\}$ in RGB frames, assigns them to detected humans, and generates gaze feature maps for each human $\{\mathbf{g}_{t,i}\}$ using a gaze-following model. Next, all features in a frame are projected by an input embedding block. The human-object pair features are concatenated to a sequence of pair representations $\mathbf{X}_t$, which are refined to $\mathbf{X}_t^\text{sp}$ by a spatial encoder. The spatial encoder also extracts a global context feature $\mathbf{c}_t$ from each frame. Then, the global features $\{\mathbf{c}_t\}$ and projected human gaze features $\{\mathbf{g}^\prime_{t,i}\}$ are concatenated to build the person-wise sliding windows of context features. Meanwhile, several instance-level sliding windows are constructed, each only containing refined pair representations of one unique human-object pair across time $\left[\mathbf{x}_{t-L+1, \langle i,j \rangle}^\text{sp},\dots,\mathbf{x}_{t, \langle i,j \rangle}^\text{sp}\right]$. A temporal encoder fuses context knowledge into the pair representations by the cross-attention mechanism. Finally, the prediction heads estimate the probability distribution $\mathbf{z}_{t,\langle i,j \rangle}$ of interactions for each human-object pair based on the last occurrence $\mathbf{x}_{t, \langle i,j \rangle}^\text{tmp}$ in the temporal encoder output. }
    \label{fig:architecture}
  \end{figure*}

\section{Our Method}
We aim to solve both HOI detection and anticipation tasks from videos with the same spatio-temporal transformer architecture. The proposed two-stage framework illustrated in Fig.~\ref{fig:architecture} is composed of an object module, a gaze module, and a spatio-temporal module. The object module and gaze module extract features from RGB frames in parallel. The spatio-temporal module based on STTran~\citep{hoi_v2:sttran} exploits these features to detect current HOIs or anticipate future HOIs. 

\subsection{Problem Setup}
Similar to the image-based HOI detection task \citep{hoi_i2:visual_semantic, hoi_i2:detecting_recognizing}, a video-based HOI detection task is defined as to retrieve bounding boxes of human subjects $\{\mathbf{b}_{t,i}^\text{s}\}$ and objects $\{\mathbf{b}_{t,j}\}$, identify object classes $\{c_{t,j}\}$, and recognize their interaction predicates $\mathbf{p}_{t,\langle i,j \rangle}$ in every frame $I_t$, where $I_t \in \mathbb{R}^{h\times w\times 3}$ denotes an RGB frame at time $t$. The subscripts $i$ and $j$ represent an arbitrary human and object. The detected HOIs are expressed as a set of triplets $\{\langle \mathbf{b}_{t,i}^\text{s}, \mathbf{p}_{t,\langle i,j \rangle}, \mathbf{b}_{t,j} \rangle\}$. 

For a video-based HOI anticipation task, we follow the setup that the model detects humans $\{\mathbf{b}_{t,i}^\text{s}\}$ and objects $\{\mathbf{b}_{t,j}\}$, $\{c_{t,j}\}$ from past observations $[I_1, \dots, I_{t}]$ and predicts HOIs $\{\langle \mathbf{b}_{t,i}^\text{s}, \mathbf{p}_{t+\tau_a,\langle i,j \rangle}, \mathbf{b}_{t,j} \rangle\}$ in the future with a fixed time gap $\tau_a$.

\subsection{Object Module}
The object module takes a sequence of $T$ RGB frames as input $V=[I_1, \dots, I_T]$. In each frame $I_t$, the object module detects $n_t$ bounding boxes $\{\mathbf{b}_{t,j}\}$, as well as the corresponding classes $\{c_{t,j}\}$. Among the $n_t$ detections, $n_t^\text{s}$ are human bounding boxes $\{\mathbf{b}_{t,i}^\text{s}\}$. An object tracker then associates current detections with past detections and obtains the trajectories of bounding boxes of human $\{\mathbf{H}_i\}$ and objects $\{\mathbf{O}_j\}$. This object tracker allows the model to analyze every unique human-object pair separately in a complex scene. After locating humans and objects in a video, it is essential to exploit features from human-object pairs to detect and anticipate the interactions. Inspired by STTran~\citep{hoi_v2:sttran}, we use a ResNet feature extractor to generate visual features $\mathbf{v}_{t,j} \in \mathbb{R}^{2048}$ for each box $\mathbf{b}_{t,j}$. The visual feature inside the subject bounding box $\mathbf{b}_{t,i}^\text{s}$ is denoted as $\mathbf{v}_{t,i}^\text{s}=\mathbf{v}_{t,i}$. In addition, leveraging the spatial relation between human and objects is crucial to recognize some actions, such as \emph{playing} or not \emph{playing} a guitar. Thus, the visual relation features $\mathbf{v}_{t,\langle i,j \rangle} \in \mathbb{R}^{2048}$ and a two-channel spatial relation binary mask $\mathbf{m}_{t,\langle i,j \rangle} \in \mathbb{R}^{2 \times 27 \times 27}$ are also generated for each human-object pair $\langle \mathbf{b}_{t,i}^\text{s}, \mathbf{b}_{t,j} \rangle$. Furthermore, possible types of interactions depend on object classes. For example, humans are more likely to \emph{ride} or \emph{carry} a bicycle than \emph{bite} a bicycle. To reflect this characteristic of HOIs, our object module uses a word embedding model~\citep{semantic:glove} to generate the object semantic feature $\mathbf{s}_{t,j} \in \mathbb{R}^{200}$ from the object category $c_{t,j}$ as an additional modality. 

\subsection{Gaze Module}

\begin{figure}
    \centering
    \includegraphics[height=120pt]{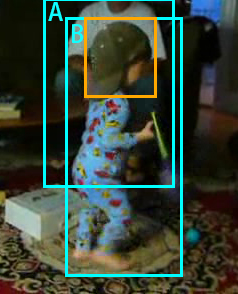}\hspace{1em}%
    \includegraphics[height=120pt]{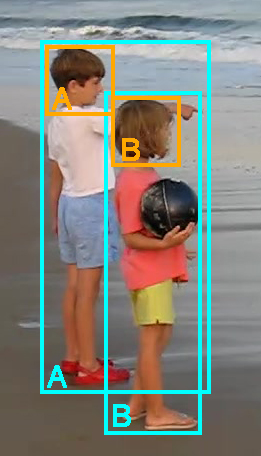}
    \caption{When detecting heads directly from human bounding boxes, there might be human-head mismatching in case when two human boxes are overlapped. In these two examples, human B's head might be mismatched to human A. Our human-head association algorithm can overcome this problem.}
    \label{fig:mismatch}
\end{figure}

We adopt the gaze-following method proposed in~\citep{gaze:detecting_attended} to generate the gaze heatmap for each human. This method requires a head image as an input. Thus, we need a head detector to identify human heads in the scene. We observe that directly obtaining the head bounding box from the human box might cause mismatches in some scenarios. As shown in both images in Fig.~\ref{fig:mismatch}, in human A's bounding box, another person's head appears. Directly obtaining head detection from human A's box may cause human B's head to be mismatched with human A. Therefore, our gaze module first retrieves $n_t^{h}$ heads $\{\mathbf{b}_{t,i}^\text{h}\}$ from the full RGB frame $I_t$. Then, all detected head bounding boxes are matched to all human bounding boxes from the object module. This process involves a linear assignment problem. We first determine which detected heads are possible matches for each human. An intersection over head (IoH) ratio is computed for every human $\mathbf{b}_{t,i}^\text{s}$ and head $\mathbf{b}_{t,k}^\text{h}$ according to Equation~\ref{eq:iou_head}, where $\mathcal{A}(\cdot)$ denotes the function for area calculation. If the IoH ratio is larger than a threshold, this head detection is considered as a shortlisted head for this human. We set the threshold to $0.7$, which allows this metric to be robust to slightly inaccurate detections. 

\begin{equation}
   \text{IoH}(i, k)=\frac{\mathcal{A}(\mathbf{b}_{t,k}^\text{h} \cap \mathbf{b}_{t,i}^\text{s})}{\mathcal{A}(\mathbf{b}_{t,k}^\text{h})}\text{,}
   \label{eq:iou_head}
\end{equation}

We apply the Jonker-Volgenant algorithm~\citep{assignment:jonker, assignment:new} to find the best human-head association for each frame. This algorithm requires a cost matrix. Intuitively, the human head is usually positioned at the limits of the body. Thus, we compute a human-head distance ratio $d_{t, \langle i, k\rangle}$ by dividing the distance between a human bounding box and a head bounding box by the length of the shorter edge of the human box. In addition, the confidence score of head detection plays an important role in the human-head association. Therefore, we use a weighted sum of the human-head distance ratio $d_{t, \langle i, k\rangle}$ and the inverse of head confidence score as the cost to assign head $\mathbf{b}_{t,k}^\text{h}$ to human $\mathbf{h}_{t,i}$. 

Finally, the gaze-following model proposed by \citet{gaze:detecting_attended} estimates human gaze heatmaps from video clips. This approach combines the head information and the scene feature map using an attention mechanism. Then, a convolutional Long Short-Term Memory (Conv-LSTM) network is applied to encode the fused features and extract temporal dependencies to estimate the gaze heatmap $\mathbf{g}_{t,i} \in \mathbb{R}^{64 \times 64}$ for each human $\mathbf{b}_{t,i}^\text{s}$ at each time step.

\subsection{Input Embedding}
At each time step $t$, the object module generates a set of features $(\mathbf{v}_{t,i}^\text{s}, \mathbf{v}_{t,j}, \mathbf{v}_{t,\langle i,j \rangle}, \mathbf{m}_{t,\langle i,j \rangle}, \mathbf{s}_{t,j})$ for the human-object pair $\langle \mathbf{b}_{t,i}^\text{s}, \mathbf{b}_{t,j} \rangle$. Meanwhile, the gaze module outputs human gaze heatmaps $\mathbf{g}_{t,i}$ for each human. To reduce the dimensionality and optimize the model efficiency, these features need to be encoded before being fed to the spatio-temporal transformer. Inspired by STTran~\citep{hoi_v2:sttran}, we use linear projection matrices $\mathbf{W}^\text{s} \in \mathbb{R}^{2048 \times 512}$ and $\mathbf{W}^\text{o} \in \mathbb{R}^{2048 \times 512}$ to compress the dimensionality of human visual features $\mathbf{v}_{t,i}^\text{s}$ and object visual features $\mathbf{v}_{t,j}$ from $2048$-d to $512$-d. The visual relation features $\mathbf{v}_{t,\langle i,j \rangle}$ are projected to $256$-d with $\mathbf{W}^\text{vr} \in \mathbb{R}^{2048 \times 256}$. To extract features from the two-channel spatial relation mask, a two-layer CNN $\mathit{f}_\text{mask}(\cdot)$ introduced in~\citep{hoi_i2:neural_motifs} with an average pooling layer at the end is applied to transform  $\mathbf{m}_{t,\langle i,j \rangle}$ to a $256$-d vector. The same CNN structure $\mathit{f}_\text{gaze}(\cdot)$ is adopted to transform the human gaze heatmap $\mathbf{g}_{t,i}$ to a $512$-d vector $\mathbf{g}^\prime_{t,i}$. The semantic feature vector $\mathbf{s}_{t,j}$ remains untouched. L2-normalization is applied to each feature vector to ensure that every feature vector has a similar data distribution. Finally, all feature vectors for the human-object pair $\langle \mathbf{b}_{t,i}^\text{s}, \mathbf{b}_{t,j} \rangle$ are concatenated to a relation representation vector $\mathbf{x}_{t,\langle i,j \rangle} \in \mathbb{R}^{1736}$. Note that all projection matrices and CNNs are jointly trained with the spatio-temporal transformer. 

\begin{figure}
    \centering
    \begin{subfigure}[b]{0.195\textwidth}
        \centering
        \begin{tikzpicture}[font=\small, >={latex}]
            \node[draw, rectangle, align=center, fill=orange!20, very thick, minimum height=20pt, minimum width=65pt] (sa) {Multi-Head\\Self-Attention};
            \node[draw, rectangle, align=center, fill=green!20, very thick, minimum height=15pt, minimum width=65pt, above=3pt of sa] (add1) {Add \& Norm};
            \node[draw, rectangle, align=center, fill=cyan!20, very thick, minimum height=20pt, minimum width=65pt, above=15pt of add1] (ffn) {Feed\\Forward};
            \node[draw, rectangle, align=center, fill=green!20, very thick, minimum height=15pt, minimum width=65pt, above=3pt of ffn] (add2) {Add \& Norm};

            \node[draw, rectangle, align=center, fill=pink!20, very thick, minimum height=15pt, minimum width=80pt, below=25pt of sa, xshift=5pt] (input) {Pair Representations};
            \node[draw, rectangle, align=center, fill=pink!20, very thick, minimum height=15pt, minimum width=5pt, left=-1pt of input] (global) {*};

            \node[draw=none, rectangle, align=center, below=8pt of global, xshift=30pt] (c) {Learned Global Token};
            \node[draw=none, rectangle, align=center, above=8pt of add2, xshift=-25pt] (c_out) {Global\\Feature $\mathbf{c}_t$};
            \node[draw=none, rectangle, align=center, above=30pt of add2, xshift=15pt] (output) {Refined Pair\\Representations\\$\mathbf{X}_t^\text{sp}$};

            \node[draw, color=gray, rectangle, align=center, fill=none, very thick, dashed, minimum height=127pt, minimum width=90pt, above=3pt of input, xshift=-10pt] (box) {};
            \node[draw=none, color=gray, rectangle, align=center, above=-2pt of box, xshift=40pt] (n) {$N_\text{sp}\times$};

            \draw[-] (sa.north) -> (add1.south);
            \draw[->] (add1.north) -> (ffn.south);
            \draw[-] (ffn.north) -> (add2.south);
            \draw[->] (add1.north) --++ (0, 6pt) --++ (-45pt, 0) |- (add2.west);
            \draw[->] ([xshift=-5pt]input.north) -> (sa.south);
            \draw[->] ([xshift=-5pt]input.north) --++ (0, 15pt) --++ (-20pt, 0) -> ([xshift=-20pt]sa.south);
            \draw[->] ([xshift=-5pt]input.north) --++ (0, 15pt) --++ (20pt, 0) -> ([xshift=20pt]sa.south);
            \draw[->] ([xshift=-5pt]input.north) --++ (0, 6pt) --++ (-45pt, 0) |- (add1.west);
            \draw[->] ([xshift=-30pt]c.north) -> (global.south);
            \draw[->] ([xshift=-25pt]add2.north) -> (c_out.south);
            \draw[->] ([xshift=15pt]add2.north) -> (output.south);
        \end{tikzpicture}
        \caption{Our spatial encoder with the global token.}
        \label{fig:spatial_encoder}
    \end{subfigure}\hspace{1em}%
    \begin{subfigure}[b]{0.26\textwidth}
        \centering
        \begin{tikzpicture}[font=\small, >={latex}]
            \node[draw, rectangle, align=center, fill=orange!20, very thick, minimum height=20pt, minimum width=65pt] (sa) {Multi-Head\\Self-Attention};
            \node[draw, rectangle, align=center, fill=green!20, very thick, minimum height=15pt, minimum width=65pt, above=3pt of sa] (add1) {Add \& Norm};
            \node[draw, rectangle, align=center, fill=orange!20, very thick, minimum height=20pt, minimum width=65pt, above=15pt of add1] (ca) {Multi-Head\\Cross-Attention};
            \node[draw, rectangle, align=center, fill=green!20, very thick, minimum height=15pt, minimum width=65pt, above=3pt of ca] (add2) {Add \& Norm};
            \node[draw, rectangle, align=center, fill=cyan!20, very thick, minimum height=20pt, minimum width=65pt, above=10pt of add2] (ffn) {Feed\\Forward};
            \node[draw, rectangle, align=center, fill=green!20, very thick, minimum height=15pt, minimum width=65pt, above=3pt of ffn] (add3) {Add \& Norm};
            \node[draw, color=gray, rectangle, align=center, fill=none, dashed, very thick, minimum height=35pt, minimum width=80pt, above=10pt of add3] (encoder) {Conventional\\Transformer\\Encoder Layers};
            
            \node[draw, circle, align=center, fill=none, inner sep=0, thick, minimum height=3pt, minimum width=3pt, below=20pt of sa] (plus1) {$+$};
            \node[draw, rectangle, align=center, fill=pink!20, very thick, minimum height=15pt, minimum width=30pt, below=3pt of plus1] (concat) {Concat};
            \node[draw=none, rectangle, align=center, below=5pt of concat, xshift=-30pt] (g) {Human Gaze\\Feature $\mathbf{g}^\prime_{t, i}$};
            \node[draw=none, rectangle, align=center, below=5pt of concat, xshift=25pt] (c) {Global\\Feature $\mathbf{c}_t$};

            \node[draw, circle, align=center, fill=none, inner sep=0, thick, minimum height=3pt, minimum width=3pt, above left=2pt and 20pt of add1] (plus2) {$+$};
            \node[draw=none, fill=none, minimum height=0, minimum width=0, above left=85pt and 20pt of plus2] (in) {};
            \node[draw, circle, align=center, fill=none, inner sep=0, thick, minimum height=10pt, minimum width=10pt, below=64pt of plus2] (pos) {$\sim$};
            \node[draw=none, rectangle, align=center, below=-1pt of pos] (posenc) {Positional\\Encoding};

            \node[draw, color=gray, rectangle, align=center, fill=none, very thick, dashed, minimum height=178pt, minimum width=90pt, above=2pt of plus1, xshift=-5pt] (box) {};
            \node[draw=none, color=gray, rectangle, align=center, above=-10pt of box, xshift=-52pt] (n) {$1\times$};
            \node[draw=none, color=gray, rectangle, align=center, above=-2pt of encoder, xshift=-30pt] (n2) {$(N_\text{tmp}-1)\times$};

            \node[draw=none, fill=none, minimum height=0, minimum width=0, above=10pt of encoder] (out) {};

            \node[draw=none, fill=none, minimum height=0, minimum width=0, below=-2pt of ca, xshift=-25pt] (q) {Q};
            \node[draw=none, fill=none, minimum height=0, minimum width=0, below=-2pt of ca, xshift=-5pt] (k) {K};
            \node[draw=none, fill=none, minimum height=0, minimum width=0, below=-2pt of ca, xshift=15pt] (v) {V};

            \draw[-] (sa.north) -> (add1.south);
            \draw[->] (add1.north) -> (ca.south);
            \draw[->] (add1.north) --++ (0, 5pt) -| ([xshift=20pt]ca.south);
            \draw[->] (add2.north) -> (ffn.south);
            \draw[->] (add2.north) --++ (0, 3pt) --++ (-45pt, 0) |- (add3.west);
            \draw[-] (ffn.north) -> (add3.south);
            \draw[->] (add3.north) -> (encoder.south);
            
            \draw[-] (concat.north) -> (plus1.south);
            \draw[->] ([xshift=5pt]g.north) |- (concat.west);
            \draw[->] (c.north) |- (concat.east);
            
            \draw[->] (in.east) --++ (10pt, 0) |- (plus2.west);
            \draw[->] (plus2.east) -| ([xshift=-20]ca.south);
            \draw[->] (plus2.north) |- (add2.west);
            \draw[->] (pos.east) -> (plus1.west);
            \draw[->] (pos.north) -> (plus2.south);

            \draw[->] (plus1.north) -> (sa.south);
            \draw[->] (plus1.north) --++ (0, 10pt) -| ([xshift=-20pt]sa.south);
            \draw[->] (plus1.north) --++ (0, 10pt) -| ([xshift=20pt]sa.south);
            \draw[->] (plus1.north) --++ (0, 6pt) --++ (-45pt, 0) |- (add1.west);

            \draw[->] (encoder.north) -> (out.south);

        \end{tikzpicture}
        \caption{Our temporal encoder with cross-attention layer. }
        \label{fig:cross_encoder}
    \end{subfigure}
    \caption{The architecture of our spatial and temporal encoders.}
    \label{fig:encoder}
\end{figure}
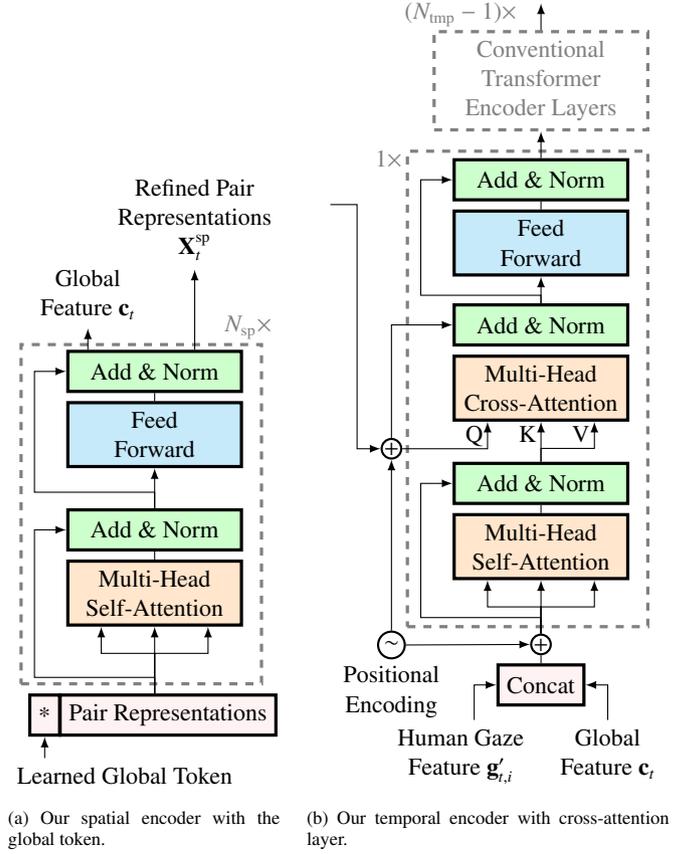

\subsection{Spatio-Temporal Module}
A spatio-temporal transformer inspired by STTran~\citep{hoi_v2:sttran} is applied to aggregate contexts from a sliding window of frames. The architecture is illustrated in Fig.~\ref{fig:encoder}. This model is composed of a spatial encoder and a temporal encoder.

First, a spatial encoder exploits human-object relation representations from one frame to understand the dependencies between the visual appearances, spatial relations, and semantic features. It also extracts a global feature vector for each frame, which is expected to represent the contexts between all human-object pairs. The spatial encoder receives the human-object pair relation representations $\mathbf{X}_t=[\mathbf{x}_{t,\langle 1,1 \rangle}, ~\dots~, \mathbf{x}_{t,\langle i,j \rangle}, \dots, \mathbf{x}_{t,\langle n_t^\text{s},n_t^\text{o} \rangle}]$ within one frame as the input. Inspired by the classification token proposed in ViT~\citep{transformer:vit}, we prepend a learnable global token to the spatial encoder input. After $N_\text{sp}$ stacked self-attention layers, the global token summarizes the dependencies between human-object pairs to a global feature vector $c_t$, while the pair relation representations are refined to $\mathbf{X}_t^\text{sp}=[\mathbf{x}^\text{sp}_{t,\langle 1,1 \rangle}, ~\dots~, \mathbf{x}^\text{sp}_{t,\langle i,j \rangle}, \dots, \mathbf{x}^\text{sp}_{t,\langle n_t^\text{s},n_t^\text{o} \rangle}]$.

Then, the refined pair representations are concatenated to several input sequences for the temporal encoder. The original STTran~\citep{hoi_v2:sttran} is designed for the Action Genome dataset~\citep{hoi_v_set:action_genome}, where only one human is annotated in each video. However, in real-world scenarios, multiple people and objects may appear. The VidHOI dataset~\citep{hoi_v_set:VidHOI} also provides full annotations for multi-person scenes. Thus, STTran may suffer from performance degradation as it treats all relation representations jointly as one sequence. In contrast, we propose to model the temporal evolution of each unique human-object pair independently. For that, we re-formulate the temporal encoder input such that each sequence only contains one particular human $i$ and object $j$, i.e., $[\mathbf{x}_{t-L+1,\langle i,j \rangle}^\text{sp}, \dots, \mathbf{x}_{t,\langle i,j \rangle}^\text{sp}]$, where $L$ denotes the length of a sliding window.

Next, referring to Fig.~\ref{fig:cross_encoder}, the gaze feature $\mathbf{g}^\prime_{t,i}$ from each unique human in a frame is concatenated with the global feature $\mathbf{c}_t$ of that frame to $\mathbf{c}_{t, i}=[\mathbf{c}_t, \mathbf{g}^\prime_{t, i}]$. The resulting vector is filled into a person-wise sliding window of high-level context features $[\mathbf{c}_{t-L+1, i}, \dots, \mathbf{c}_{t, i}]$, which are fed to the temporal encoder along with the pair-wise sliding windows. Since the temporal encoder processes all entries in a sequence in parallel, the temporal order of the entries is lost. Therefore, a positional encoding is added to all entries in both high-level context sliding window and relation representation sliding window. STTran~\citep{hoi_v2:sttran} applies a learned positional encoding, however, we observe that the sinusoidal encoding performs better in our model.

The temporal encoder fuses the high-level context features and the refined pair representations by cross-attention layers and captures the evolution of their dependencies in time, which is essential to detect and anticipate temporal-related HOIs such as \emph{push} and \emph{pull}, for instance. In the first temporal encoder layer as shown in Fig.~\ref{fig:cross_encoder}, a multi-head self-attention layer first captures temporal dependencies between high-level context features. A cross-attention layer then fuses the human-object pair representations with the high-level contexts. Same as in the vanilla transformer~\citep{transformer:vanilla}, the cross-attention is computed as: 
\begin{equation}
    \text{Attention}(Q, K, V)=\text{softmax}\left(\dfrac{QK^T}{\sqrt{d_k}}\right)V\text{.}
    \label{eq:attn}
\end{equation}
Where Q, K, and V denote queries, keys, and values. $d_k$ is the dimensionality of the keys. In our case, the queries are the pair representations and the keys and values are the high-level contexts features. The outputs of the first temporal encoder layer are fed to $N_\text{tmp}-1$ stacked conventional self-attention layers to aggregate deeper temporal dependencies between the fused features. To ensure causality, the last temporal encoder layer only outputs the representation vectors for the last frame in each sliding window, i.e., $\mathbf{x}_{t,\langle i,j \rangle}^\text{tmp}$.

Finally, a set of prediction heads generate the probability distributions for different interaction categories. Each prediction head is a one-layer feed-forward network followed by a Softmax or Sigmoid function depending on whether the classification is single-label or multi-label. The outputs of all prediction heads are concatenated to the final model output $\mathbf{z}_{t,\langle i,j \rangle}$. On the VidHOI dataset~\citep{hoi_v_set:VidHOI}, we have a spatial relation head and an action head, each with Sigmoid function. On the Action Genome~\citep{hoi_v_set:action_genome} dataset, there are three prediction heads: attention head, spatial relation head, and action head. The attention head determines whether the human is watching an object, thus is with Softmax function. The other two heads are with Sigmoid function. 

\subsection{Loss Function}
Since a human-object pair in the VidHOI dataset~\citep{hoi_v_set:VidHOI} may be labeled by multiple interactions at the same time, such as $\langle$human, \emph{next to} \& \emph{watch} \& \emph{hold}, cup$\rangle$, HOI detection and anticipation on VidHOI dataset leads to a multi-class multi-label classification problem. Binary cross-entropy (BCE) loss is usually applied in such tasks, which computes the loss for each interaction class independently to other classes. However, VidHOI dataset is an unbalanced dataset with long-tailed interaction distribution. To address the imbalance issue and avoid over-emphasizing the importance of the most frequent classes in the dataset, we adopt the class-balanced (CB) Focal loss~\citep{loss:cb} as follows: 
\begin{equation}
    \begin{aligned}
        & \text{CB}_\text{focal}(p_i, y_i) = -\frac{1-\beta}{1-\beta^{n_i}}(1-p_{y_i})^\gamma\log(p_{y_i})\text{,} \\
        & \text{with } p_{y_i}=\left\{ 
            \begin{aligned}
                & p_i    && \text{if }y_i=1 \\
                & 1-p_i  && \text{otherwise}\text{.} \\
            \end{aligned} \right. \\
        \end{aligned}
    \label{eq:cb_focal_loss}
\end{equation}
The term $-(1-p_{y_i})^\gamma\log(p_{y_i})$ refers to the Focal loss proposed in~\citep{loss:focal}, where $p_i$ denotes the estimated probability for the $i$-th class and $y_i \in \{0, 1\}$ is the ground-truth label. The variable $n_i$ denotes the number of samples in the ground truth of the $i$-th class and $\beta \in [0, 1)$ is a tunable parameter. The mean of losses in all classes is considered as the loss for one prediction.

\section{Experiments}
\subsection{Dataset and Baselines}
\subsubsection{VidHOI dataset}
We validate our framework on VidHOI dataset~\citep{hoi_v_set:VidHOI} as this is currently the largest video dataset with complete HOI annotations. The VidHOI dataset contains videos retrieved from social media where humans are performing daily activities without pre-defined scripts in highly unstructured and noisy environments. Thus, these videos represent real-world scenes. The VidHOI dataset applies keyframe-based annotations, where the keyframes are sampled in $1$ frame per second (FPS). There are $78$ object categories and $50$ predicate classes. Among the predicate classes, we define $8$ predicates as spatial relations (\emph{away}, \emph{towards}, \emph{above}, \emph{next to}, \emph{behind}, \emph{in front of}, \emph{inside}, \emph{beneath}), while the rest $42$ predicates are actions (e.g., \emph{hold}, \emph{push}, \dots). 

The ST-HOI baseline~\citep{hoi_v_set:VidHOI} is adopted as the baseline for HOI detection task on VidHOI dataset. This method extracts visual features from object trajectories by a SlowFast~\citep{cnn:slowfast} backbone and generates pose features using a spatio-temporal pose module. These features are concatenated and fed to a two-layer prediction head. In addition, we use the original STTran~\citep{hoi_v2:sttran} as another baseline model. This model is trained with the same learning rate scheduler as our model but only for $10$ epochs as suggested in their source code. The TUTOR model~\citep{hoi_v2:tubelet_tokens} is also validated on VidHOI dataset. We use their provided results for comparison.

\subsubsection{Action Genome dataset}
Action Genome~\citep{hoi_v_set:action_genome} is another large-scale video dataset containing $35$ object categories and $25$ interaction classes. Nevertheless, only HOIs for a single person are annotated in each video even if more people show up. Moreover, the videos are generated by volunteers performing pre-defined tasks. Thus, models designed on the Action Genome dataset may be less useful in the real world. We only conduct an experiment on this dataset in the HOI detection task to demonstrate the robustness of our framework.

We apply the original STTran~\citep{hoi_v2:sttran} as the baseline model on the Action Genome dataset. In addition, several image-based HOI detection models~\citep{hoi_i2:language_prior, hoi_i2:msdn, hoi_i2:vctree, hoi_i2:reidn, hoi_i2:gpsnet} are chosen for further comparison. The results of these works are provided by~\citep{hoi_v2:sttran}.  

\subsection{Evaluation Metrics}
Following the standard procedure in HOI detection, mean average precision (mAP) is adopted as one of our evaluation metrics. The mAP is a summary of precision-recall curves for all interaction classes. A predicted HOI triplet is assigned true positive if: (1) both detected human and object bounding boxes are overlapped with the ground truth with intersection over union (IoU) $>0.5$, (2) the predicted object class is correct, and (3) the predicted interaction is correct. The metric mAP is reported on the VidHOI dataset over three different HOI category sets: (1) Full: all $557$ HOI triplet categories, (2) Rare: $315$ categories with $<25$ instances in the validation set, and (3) Non-rare: $242$ categories with $\geq 25$ instances in the validation set. We apply the mAP computation method from QPIC~\citep{hoi_i1:qpic}.



For the HOI anticipation task, the mAP does not well represent the model performance as it is evaluated on all predicted HOIs in a frame. Applications of HOI anticipation usually consider the top predictions for each human separately. For example, a robot may decide how to assist a human based on the most likely HOI forecasted. On the egocentric action anticipation benchmarks~\citep{ego_set:epic100, ego_set:epic55, ego_set:egteagaze}, top-$5$ recall or top-$5$ accuracy are often employed to address such application scenarios. The egocentric videos only contain one person as the subject, and only one action is performed in each frame. Thus, evaluating the top-$k$ predictions in one frame is equivalent to evaluating the top-$k$ predictions for one human. Inspired by this idea, we propose a set of person-wise multi-label top-$k$ metrics as additional evaluation metrics. For each frame, we first assign the detected human-object pairs to the ground-truth pairs. Then, the top-$k$ triplets of each human are used to compute the metrics for this human. We follow~\citep{metric:multi_label} to calculate the multi-label recall, precision, accuracy, and F1-score. On the VidHOI dataset, we report the person-wise multi-label top-$k$ metrics with $k=5$ and confidence threshold $=0.3$. The final results are averaged over all humans in the dataset, without frame-wise or video-wise mean computation. On the Action Genome dataset, most baselines only consider the Recall@$k$ metric, which is identical to person-wise top-$k$ recall since Action Genome only consists of single-person scenes. The final results are averaged frame-wise. 

All models are trained with ground-truth object trajectories. We follow the two evaluation modes defined in ST-HOI baseline~\citep{hoi_v_set:VidHOI}: models in \emph{Oracle} mode are evaluated with ground-truth object bounding boxes, while models in \emph{Detection} mode are evaluated with object detector. During the evaluation in \emph{Detection} mode, the ST-HOI baseline~\citep{hoi_v_set:VidHOI} removes the frames without any object detected. This trick could increase the recall as some not detected ground-truth HOIs are filtered out. We use their reported mAP value for comparison, but we evaluate our model without excluding any frames. In the frames with no valid object detection, all ground-truth HOIs are regarded as false negatives.

By observing a sequence of past $T$ frames, the model is expected to detect HOIs in the last observed frame (detection task) or forecast HOIs in the $\tau_a$-th future frame (anticipation task). For the anticipation task, we train and validate our models with $\tau_a \in \{1, 3, 5, 7\}$, where for example, $\tau_a=5$ means $5$ seconds in the future in VidHOI dataset. The anticipation times are intuitively selected to show the performance of HOI anticipation in the near future. The evaluations for the anticipation task are only conducted on those videos that are enough long for $\tau_a=7$. A potential issue in HOI anticipation task in third-person videos is that the humans and objects in the current frame may disappear in the future due to the movement of humans or the camera. Thus, for mAP computation, we ignore the anticipations that are matched to a ground-truth human-object pair which is not available in the future. For our proposed person-wise top-$k$ metrics, the persons out of frame in the future are excluded.

    

\subsection{Implementation Details}
For our object module, we employ YOLOv5 model~\citep{detection:yolov5} as the object detector. The weights are pre-trained on COCO dataset~\citep{detection:COCO} and finetuned for the VidHOI dataset. We apply the pre-trained DeepSORT model~\citep{tracking:deepsort} as the human tracker, ResNet-101~\citep{cnn:ResNet} as feature backbone, and GloVe model~\citep{semantic:glove} for word embedding. 

In the gaze module, we also apply YOLOv5 to detect heads from RGB frames. The model is pre-trained on the Crowdhuman dataset~\citep{detection:crowdhuman}. The gaze-following method introduced in~\citep{gaze:detecting_attended} and pre-trained on the VideoAttentionTarget dataset~\citep{gaze:detecting_attended} is adopted to generate gaze features. All weights in the object module and gaze module are frozen during the training of the spatio-temporal transformer. 

The training procedure from STTran~\citep{hoi_v2:sttran} has a limitation that it collapses to overfitting quickly as it samples a batch of windows from the same video at each training step. To tackle this issue, we design a new data sampling strategy to sample a batch of windows from different videos, and each video is only visited once in an epoch. In addition, we introduce random horizontal flipping as data augmentation. The hyperparameters of our model are finetuned on the VidHOI dataset. For the experiment on the Action Genome dataset, we simply reuse the same setup as on the VidHOI dataset. 

Following the original STTran~\citep{hoi_v2:sttran}, our spatio-temporal transformer model has $2048$-d FFN layers and $8$ heads in multi-head attention layers. The spatial encoder consists of $1$ layer while the temporal encoder contains $3$ layers. The sliding window length is set to $6$ according to the ablation study. We adopt CB Focal loss with $\gamma=0.5$ and $\beta=0.9999$ which are recommended for large-scale and extremely imbalanced datasets in~\citep{loss:cb}. Mini-batch learning is used to accelerate the training. We train the model using AdamW optimizer~\citep{optimizer:adamw} with $3$ warming-up epochs with an initial learning rate of $1 \times 10^{-8}$, a peak learning rate of $1 \times 10^{-4}$, and an exponential decay with factor $0.1$. The weight decay factor is set to $1 \times 10^{-2}$ and the dropout rate is $0.1$. All trainings are run for $25$ epochs. For reproducibility, we set a fixed random seed for all training. The experiments are performed on a single NVIDIA RTX 4090 GPU.

\subsection{Quantitative Results}

\begin{table}
    \centering
    \begin{subtable}[b]{0.48\textwidth}
        \centering
        \small
        \begin{tabular}{|c|c c c|}
            \hline
            \multirow{2}{*}{Method} & \multicolumn{3}{c|}{mAP}\\
            \cline{2-4}
            & Full & Non-rare & Rare\\
            \hhline{|=|= = =|}
            GPNN~\citep{hoi_v2:learning_gpn} & 18.47 & 24.50 & 16.41\\
            STIGPN~\citep{hoi_v2:st_gpn} & 19.39 & 28.13 & 18.22\\
            ST-HOI~\citep{hoi_v_set:VidHOI} & 17.60 & 27.20 & 17.30\\
            \hline
            HOTR~\citep{hoi_i1:hotr} & 21.14 & 30.75 & 19.83\\
            QPIC~\citep{hoi_i1:qpic} & 21.40 & 32.90 & 20.56\\
            TUTOR~\citep{hoi_v2:tubelet_tokens} & 26.92 & 37.12 & 23.49\\
            \hline
            STTran~\citep{hoi_v2:sttran} & 28.32 & 42.08 & 17.74\\
            Ours & \textbf{38.61} & \textbf{52.44} & \textbf{27.99}\\
            \hline
        \end{tabular}
        \caption{HOI \textbf{detection} in \textit{Oracle} mode on VidHOI validation set.}
        \label{table:detection_oracle}
    \end{subtable}

    \begin{subtable}[b]{0.48\textwidth}
        \centering
        \small
        \begin{tabular}{|c|c|c c c|}
            \hline
            \multirow{2}{*}{Method} & Object & \multicolumn{3}{c|}{mAP}\\
            \cline{3-5}
            & Detector & Full & Non-rare & Rare\\
            \hhline{|=|=|= = =|}
            ST-HOI & Detectron2 & 3.10 & 5.90 & 2.10\\
            STTran & YOLOv5 & 7.61 & 13.18 & 3.33\\
            Ours & Detectron2 & 8.83 & 14.47 & 4.50\\
            Ours & YOLOv5 & \textbf{10.40} & \textbf{16.83} & \textbf{5.46}\\
            \hline
        \end{tabular}
        \caption{HOI \textbf{detection} in \textit{Detection} mode on VidHOI validation set.}
        \label{table:detection_detection}
    \end{subtable}

    \caption{Experimental results in \textbf{HOI detection} task on VidHOI dataset~\citep{hoi_v_set:VidHOI}. The bold numbers indicate the best scores. The mAP is reported in Full, None-rare, and Rare splits. For \textit{Oracle} mode, the results of all baselines except STTran~\citep{hoi_v2:sttran} are cited from~\citep{hoi_v2:tubelet_tokens}. For \textit{Detection} mode, only the evaluation result from ST-HOI baseline is available.}
    \label{table:detection}
\end{table}

\begin{table}
    \centering
    \small
    \begin{tabular}{|c|c c c|}
       \hline
       Method & Rec@10 & Rec@20 & Rec@50 \\
       \hhline{|=|= = =|}
       VRD~\citep{hoi_i2:language_prior} & 55.5 & 64.9 & 65.2 \\
       MSDN~\citep{hoi_i2:msdn} & 69.6 & 78.9 & 79.9 \\
       VCTREE~\citep{hoi_i2:vctree} & 70.1 & 78.2 & 79.6 \\
       ReIDN~\citep{hoi_i2:reidn} & 70.7 & 78.8 & 80.3 \\
       GPS-Net~\citep{hoi_i2:gpsnet} & 71.3 & 81.2 & 82.0 \\
       STTran~\citep{hoi_v2:sttran} & 73.2 & 83.1 & 84.0 \\
       Ours & \textbf{75.4} & \textbf{83.7} & \textbf{84.3} \\ 
       \hline
    \end{tabular}

    \caption{Experimental results in \textbf{HOI detection} task on the Action Genome dataset~\citep{hoi_v_set:action_genome}. The models are evaluated in \emph{Oracle} (also called \emph{PredCLS}) mode and \emph{Semi Constraint} setup. The scores of other models are cited from~\citep{hoi_v2:sttran}. }
    \label{table:ag}
\end{table}

\begin{table}
    \centering
    \begin{subtable}[]{0.47\textwidth}
        \centering
        \small
        \begin{tabular}{|c|c|c|c c c c|}
            \hline
            \multirow{2}{*}{Method} & \multirow{2}{*}{$\tau_a$} & mAP & \multicolumn{4}{c|}{Person-wise top-5} \\
            \cline{3-7}
            & & Full & Rec & Prec & Acc & F1 \\
            \hhline{|=|=|=|= = = =|}
            \multirow{4}{*}{STTran} & 1 & 29.09 & \textbf{74.76} & 41.36 & 36.61 & 50.48\\
            & 3 & 27.59 & \textbf{74.79} & 40.86 & 36.42 & 50.16\\
            & 5 & 27.32 & \textbf{75.65} & 41.18 & 36.92 & 50.66\\
            & 7 & 26.26 & \textbf{75.69} & 40.42 & 36.27 & 50.08\\
            \hline
            \multirow{4}{*}{Ours} & 1 & \textbf{37.59} & 72.17 & \textbf{59.98} & \textbf{51.65} & \textbf{62.78}\\
            & 3 & \textbf{33.14} & 71.88 & \textbf{60.44} & \textbf{52.08} & \textbf{62.87}\\
            & 5 & \textbf{32.75} & 71.25 & \textbf{59.09} & \textbf{51.14} & \textbf{61.92}\\
            & 7 & \textbf{31.70} & 70.48 & \textbf{58.80} & \textbf{50.56} & \textbf{61.36}\\
            \hline
        \end{tabular}
        \caption{HOI \textbf{anticipation} in \emph{Oracle} mode on VidHOI validation set.}
        \label{table:anticipation_oracle}
    \end{subtable}

    \begin{subtable}[]{0.47\textwidth}
        \centering
        \small
        \begin{tabular}{|c|c|c|c c c c|}
            \hline
            \multirow{2}{*}{Method} & \multirow{2}{*}{$\tau_a$} & mAP & \multicolumn{4}{c|}{Person-wise top-5} \\
            \cline{3-7}
            & & Full & Rec & Prec & Acc & F1 \\
            \hline
            \multirow{4}{*}{STTran} & 1 & 8.80 & \textbf{53.31} & 27.62 & 18.85 & 27.15\\
            & 3 & 8.32 & \textbf{52.58} & 26.99 & 18.41 & 26.48\\
            & 5 & 8.67 & \textbf{52.96} & 26.97 & 18.48 & 26.54\\
            & 7 & 8.75 & \textbf{52.18} & 26.35 & 18.01 & 25.90\\
            \hline
            \multirow{4}{*}{Ours} & 1 & \textbf{11.30} & 52.53 & \textbf{43.61} & \textbf{28.81} & \textbf{35.86}\\
            & 3 & \textbf{10.65} & 51.63 & \textbf{43.60} & \textbf{28.66} & \textbf{35.37}\\
            & 5 & \textbf{10.19} & 51.69 & \textbf{42.49} & \textbf{28.22} & \textbf{34.88}\\
            & 7 & \textbf{10.14} & 50.72 & \textbf{42.10} & \textbf{27.60} & \textbf{34.14}\\
            \hline
        \end{tabular}
        \caption{HOI \textbf{anticipation} with YOLOv5 on VidHOI validation set.}
        \label{table:anticipation_det}
    \end{subtable}

    \caption{Experimental results in \textbf{HOI anticipation} task on the VidHOI dataset~\citep{hoi_v_set:VidHOI}. Only the mAP Full is shown as the HOI category split varies for different $\tau_a$. }
    \label{table:anticipation}
\end{table}

Table~\ref{table:detection} shows the experimental results of baselines and our framework in the HOI detection task on the VidHOI dataset. In \emph{Oracle} mode, our model consistently outperforms all recent baselines. Moreover, our extensions to the STTran~\citep{hoi_v2:sttran} lead to a significant performance boost. In \emph{Detection} mode, we additionally validate our model with the object traces generated by ST-HOI baseline using Detectron2~\citep{detection:detectron2}. The results imply that the quality of the object detector plays an important role in two-stage HOI detectors and our adopted YOLOv5 model is superior to Detectron2 in this case. However, the critical performance gap between the \textit{Oracle} mode and \textit{Detection} mode indicates that the object detector still has a large space for improvement.

The experimental results in the HOI detection task on the Action Genome dataset are listed in Table~\ref{table:ag}. We only evaluate our model in \emph{Oracle} mode (or also called \emph{PredCLS} in the baseline approaches) and \emph{Semi Constraint} setup, where \emph{Semi Constraint} means all HOI predictions with a confidence score higher than a threshold are regarded as positives. Even without a specific hyperparameter finetuning, our model still outperforms all baselines in all Recall@$k$ metrics. These results indicate the robustness of our model.  

\begin{figure*}
    \centering
    \begin{subfigure}[]{0.24\textwidth}
        \includegraphics[width=0.99\textwidth]{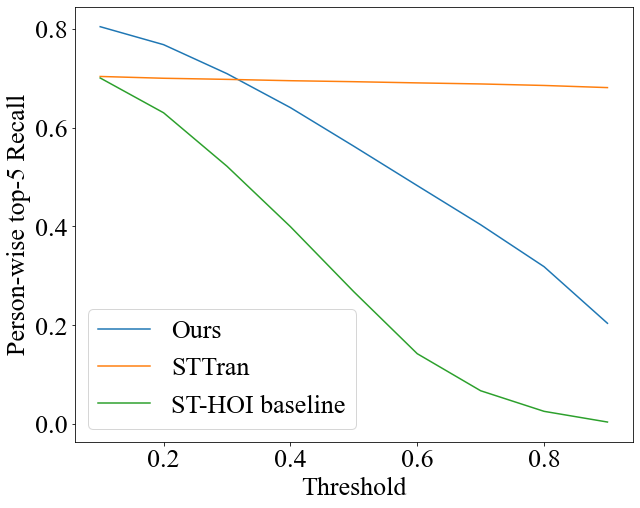}
        \caption{Person-wise top-5 Recall}
    \end{subfigure}
    \begin{subfigure}[]{0.24\textwidth}
        \includegraphics[width=0.99\textwidth]{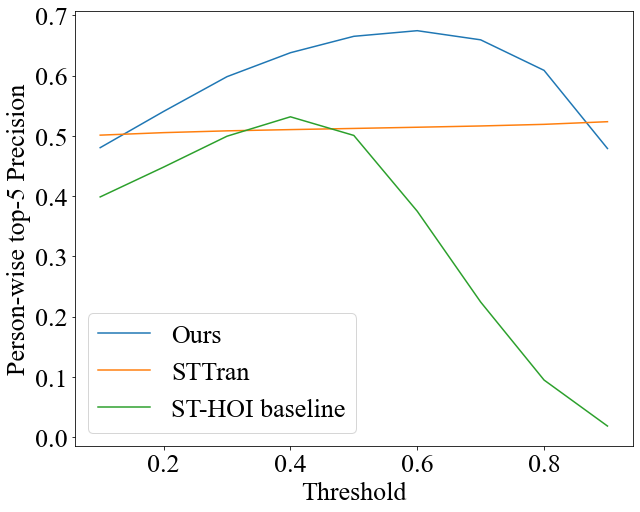}
        \caption{Person-wise top-5 Precision}
    \end{subfigure}
    \begin{subfigure}[]{0.24\textwidth}
        \includegraphics[width=0.99\textwidth]{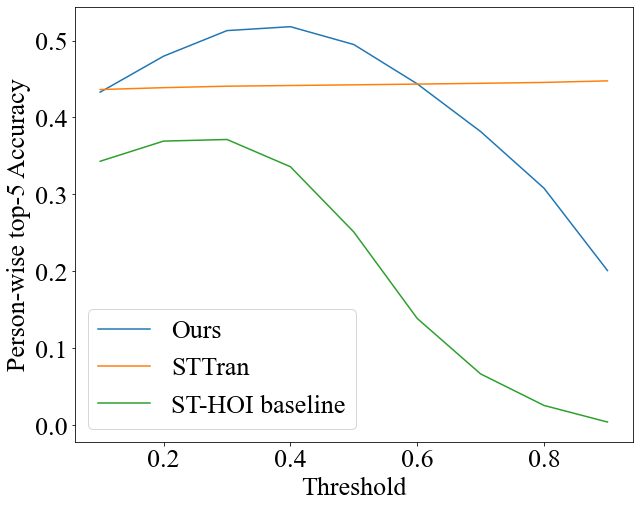}
        \caption{Person-wise top-5 Accuracy}
    \end{subfigure}
    \begin{subfigure}[]{0.24\textwidth}
        \includegraphics[width=0.99\textwidth]{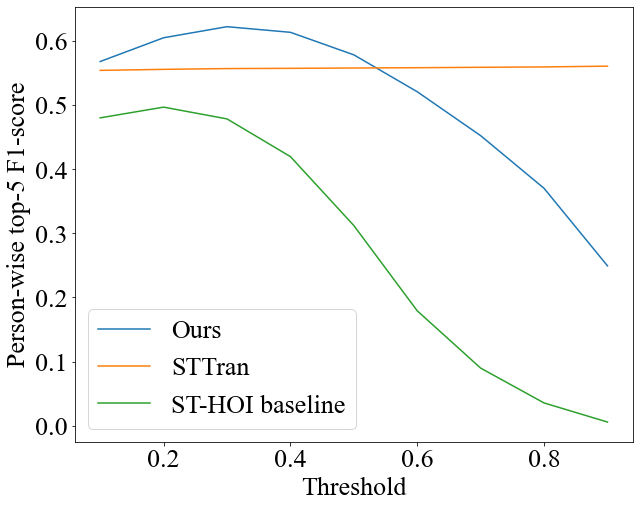}
        \caption{Person-wise top-5 F1-score}
    \end{subfigure}

    \caption{Person-wise top-$5$ metrics-threshold curves for HOI \textbf{detection} task in \textit{Oracle} mode. }
    \label{fig:ablation_thres}
\end{figure*}

The quantitative results in the HOI anticipation task are reported in Table~\ref{table:anticipation}. The non-rare and rare splits for mAP are not applicable as some ground-truth triplets are not available for anticipation due to too short videos or invisible future human-object pairs. Our model outperforms the STTran~\citep{hoi_v2:sttran} by a great margin in all metrics except the person-wise top-$5$ recall. The reason for this phenomenon is that the recall value highly depends on the confidence threshold. We additionally plot the person-wise top-$5$ score-threshold curves in Figure~\ref{fig:ablation_thres}. According to these curves, we set $0.3$ as the threshold for our model, which corresponds to the peak of accuracy and F1-score. With this threshold, our model achieves a slightly lower recall than the STTran baseline but much higher precision, accuracy, and F1-score. If a higher recall is preferred, we can shift the threshold to $0.2$, where our model beats the baseline in all metrics, but the average gain drops. 

In addition, we test the average inference time of each module in our framework. The object module with YOLOv5 object detector and DeepSORT tracker can operate at $61.3$ FPS, while the gaze module with YOLOv5 head detector and the gaze following model Chong \etal~\citep{gaze:detecting_attended} runs at $48.1$ FPS. Our proposed spatio-temporal transformer consists of $147.7$M parameters and can process $9.7$ sliding windows per second, which is real-time capable for the VidHOI dataset with a sample rate of $1$ FPS. In comparison, the original transformer in STTran~\citep{hoi_v2:sttran} contains $124.8$M parameters and achieves $25.1$ windows per second inference speed. The main reason for this speed difference lies in that the sliding window length $L$ in STTran is $2$, whereas our model has $L=6$. If we also set the sliding window length to $2$, we can achieve $23.8$ windows per second inference speed. In this setup, our model performs slightly worse with the mAP Full of $37.50$, which is still much higher than STTran. Moreover, for real applications, our model's inference time can be reduced by using a buffer to store the spatial encoder output for consecutive windows. 

\subsection{Qualitative Results}

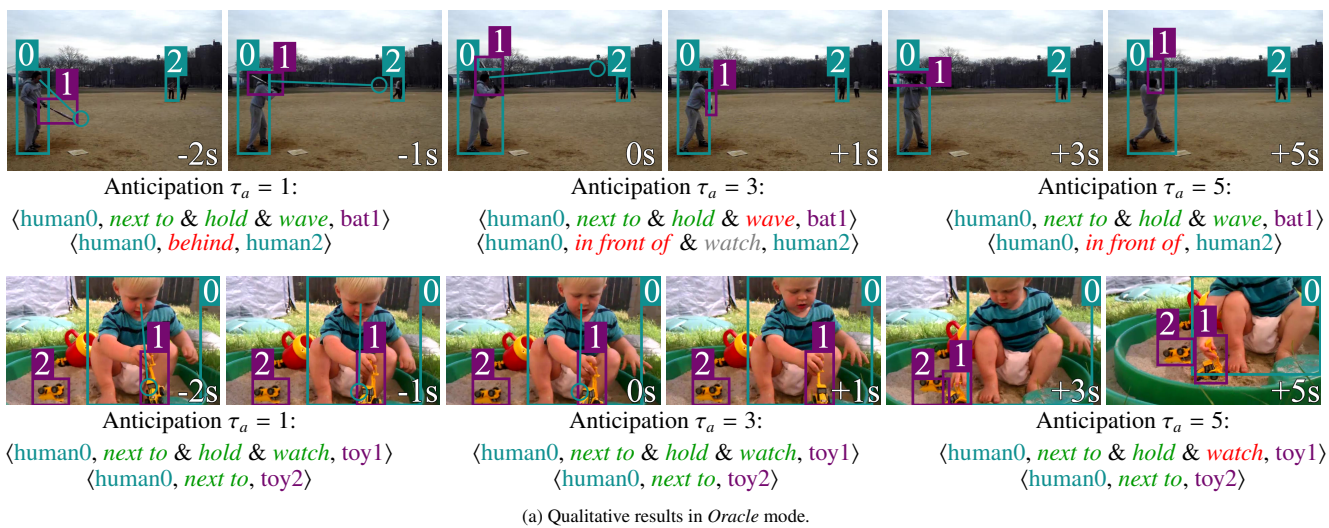
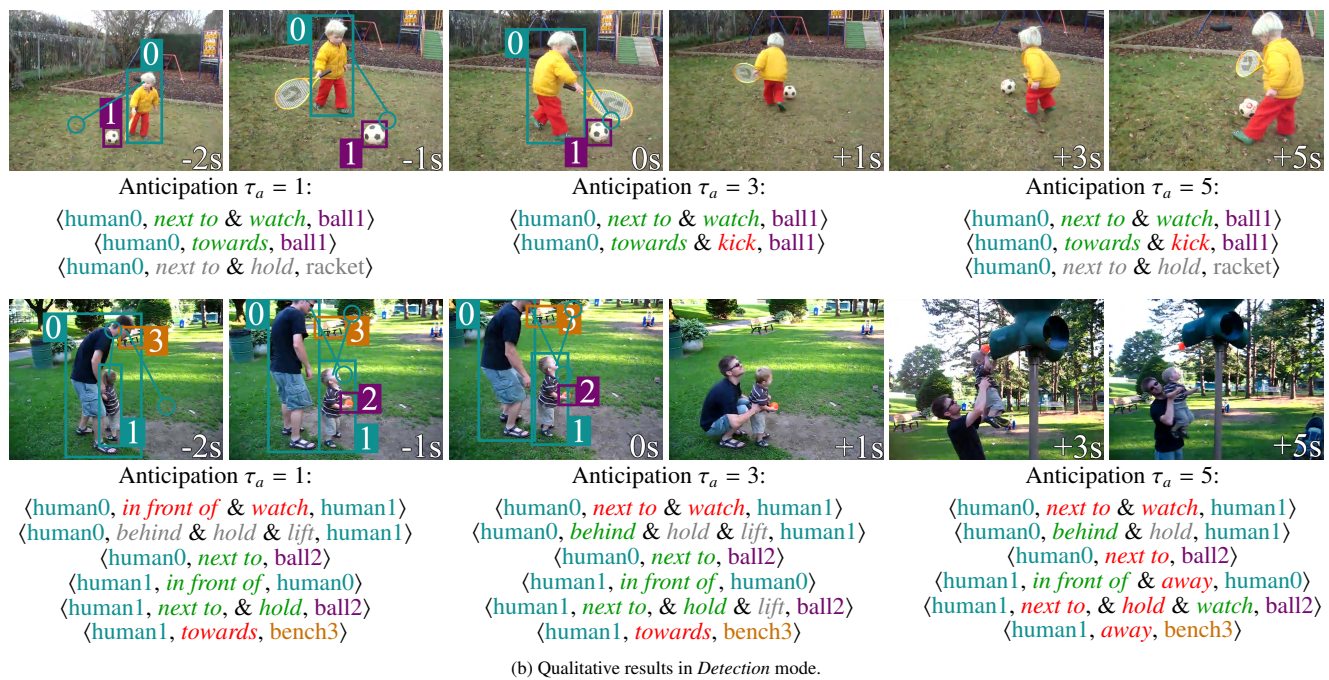
\begin{figure*}
    \centering
    \begin{subfigure}[]{0.999\textwidth}
        \centering
        \begin{tikzpicture}[font=\small, >={latex}]
            \node[inner sep=0pt] (f__2) {\includegraphics[width=80pt]{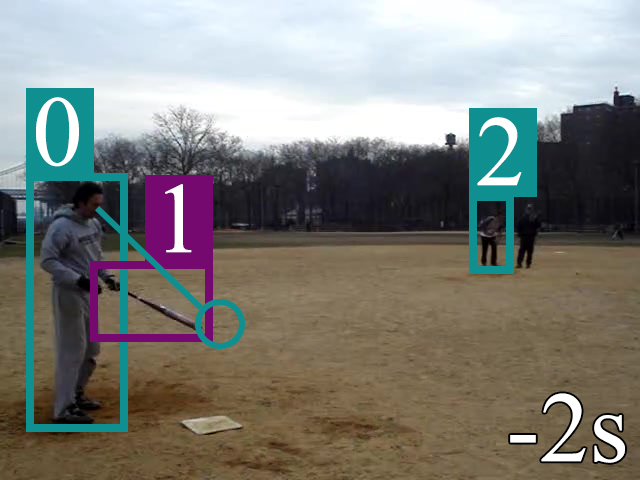}};
            \node[inner sep=0pt, right=2pt of f__2] (f__1) {\includegraphics[width=80pt]{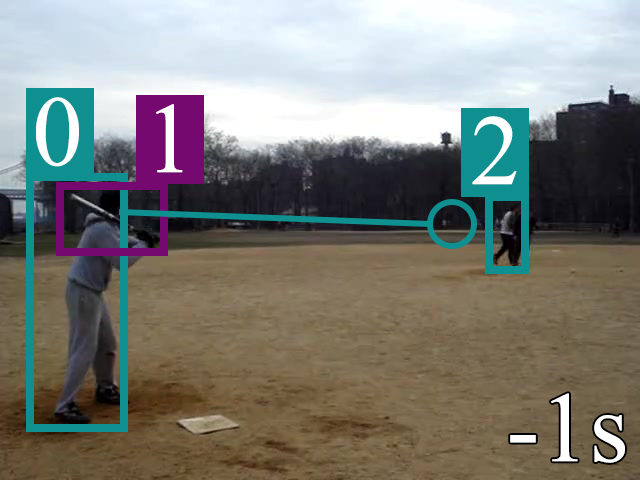}};
            \node[inner sep=0pt, right=2pt of f__1] (f_0) {\includegraphics[width=80pt]{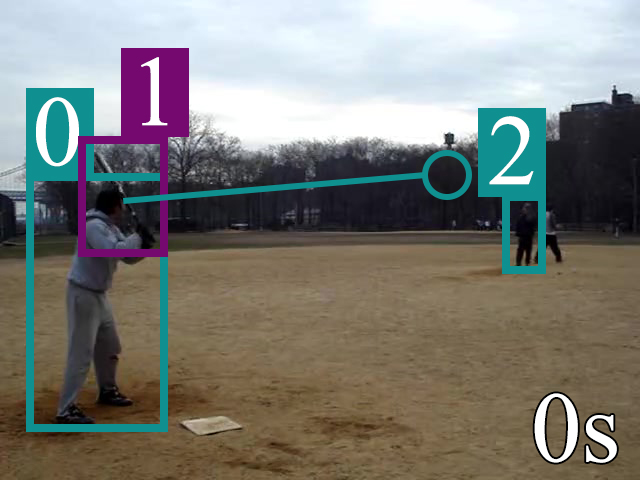}};
            \node[inner sep=0pt, right=2pt of f_0] (f_1) {\includegraphics[width=80pt]{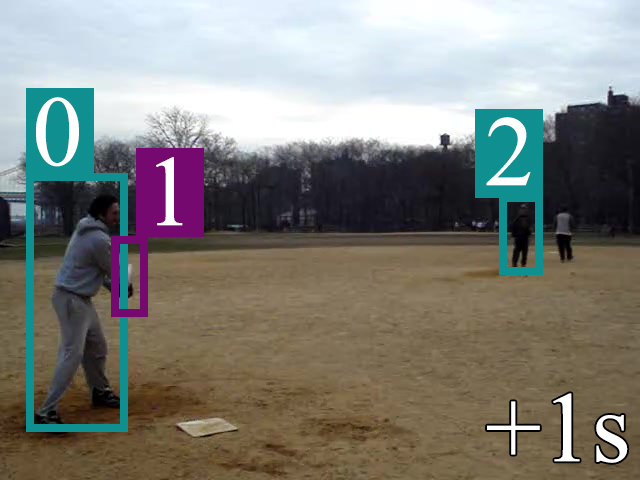}};
            \node[inner sep=0pt, right=2pt of f_1] (f_3) {\includegraphics[width=80pt]{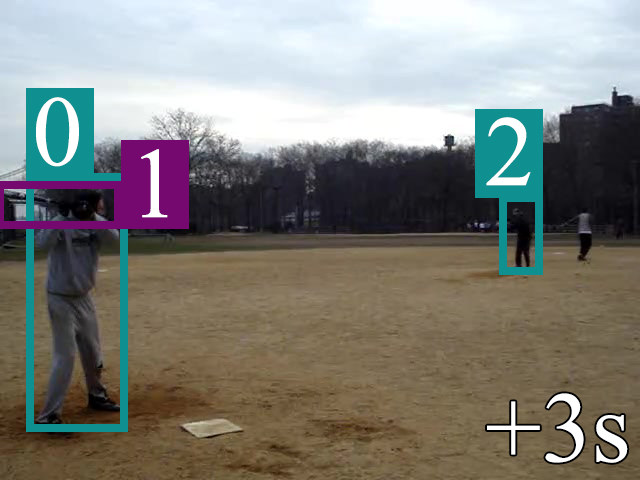}};
            \node[inner sep=0pt, right=2pt of f_3] (f_5) {\includegraphics[width=80pt]{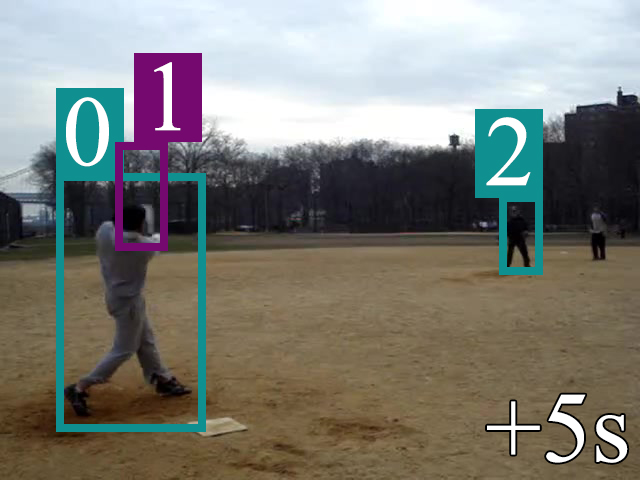}};
    
            \node[draw=none, rectangle, align=center, below left=-1pt and -30pt of f__1] (a1) {Anticipation $\tau_a=1$:};
            \node[draw=none, inner sep=0pt, rectangle, align=center, below=0pt of a1] (a11) {$\langle$\textcolor{darkcyan}{human0}, \textcolor{darkgreen}{\emph{next to}} \& \textcolor{darkgreen}{\emph{hold}} \& \textcolor{darkgreen}{\emph{wave}}, \textcolor{darkpurple}{bat1}$\rangle$};
            \node[draw=none, inner sep=0pt, rectangle, align=center, below=1pt of a11] (a12) {$\langle$\textcolor{darkcyan}{human0}, \textcolor{red}{\emph{behind}}, \textcolor{darkcyan}{human2}$\rangle$};
    
            \node[draw=none, rectangle, align=center, below left=-1pt and -40pt of f_1] (a3) {Anticipation $\tau_a=3$:};
            \node[draw=none, inner sep=0pt, rectangle, align=center, below=0pt of a3] (a31) {$\langle$\textcolor{darkcyan}{human0}, \textcolor{darkgreen}{\emph{next to}} \& \textcolor{darkgreen}{\emph{hold}} \& \textcolor{red}{\emph{wave}}, \textcolor{darkpurple}{bat1}$\rangle$};
            \node[draw=none, inner sep=0pt, rectangle, align=center, below=1pt of a31] (a32) {$\langle$\textcolor{darkcyan}{human0}, \textcolor{red}{\emph{in front of}} \& \textcolor{gray}{\emph{watch}}, \textcolor{darkcyan}{human2}$\rangle$};
    
            \node[draw=none, rectangle, align=center, below left=-1pt and -50pt of f_5] (a5) {Anticipation $\tau_a=5$:};
            \node[draw=none, inner sep=0pt, rectangle, align=center, below=0pt of a5] (a51) {$\langle$\textcolor{darkcyan}{human0}, \textcolor{darkgreen}{\emph{next to}} \& \textcolor{darkgreen}{\emph{hold}} \& \textcolor{darkgreen}{\emph{wave}}, \textcolor{darkpurple}{bat1}$\rangle$};
            \node[draw=none, inner sep=0pt, rectangle, align=center, below=1pt of a51] (a52) {$\langle$\textcolor{darkcyan}{human0}, \textcolor{red}{\emph{in front of}}, \textcolor{darkcyan}{human2}$\rangle$};
    
            \node[draw=none, rectangle, align=center, minimum height=1pt, minimum width=1pt, below=0pt of a32] (placeholder) {};
            
        \end{tikzpicture}
    
        \begin{tikzpicture}[font=\small, >={latex}]
            \node[inner sep=0pt] (f__2) {\includegraphics[width=80pt]{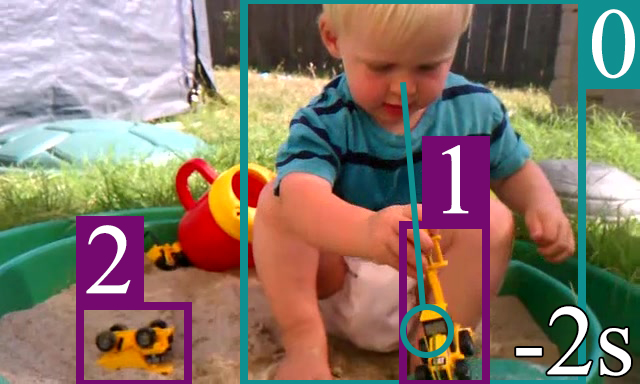}};
            \node[inner sep=0pt, right=2pt of f__2] (f__1) {\includegraphics[width=80pt]{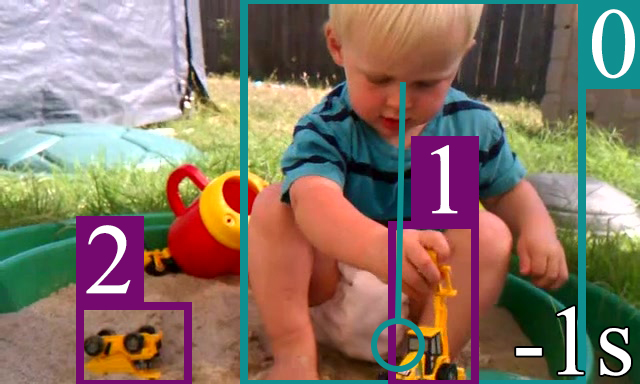}};
            \node[inner sep=0pt, right=2pt of f__1] (f_0) {\includegraphics[width=80pt]{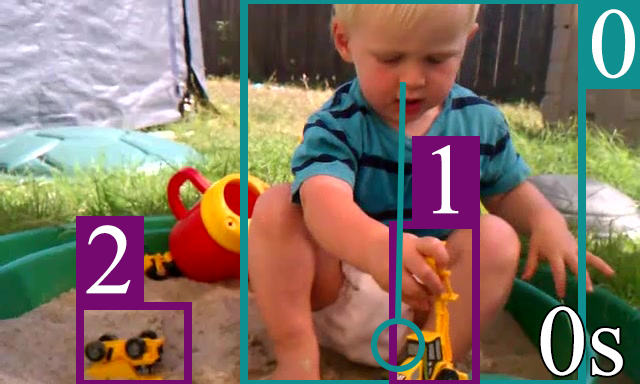}};
            \node[inner sep=0pt, right=2pt of f_0] (f_1) {\includegraphics[width=80pt]{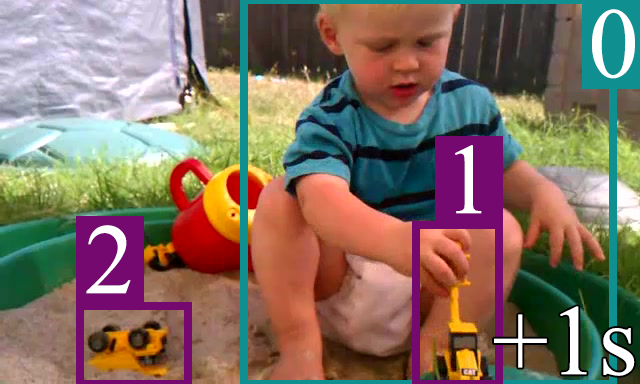}};
            \node[inner sep=0pt, right=2pt of f_1] (f_3) {\includegraphics[width=80pt]{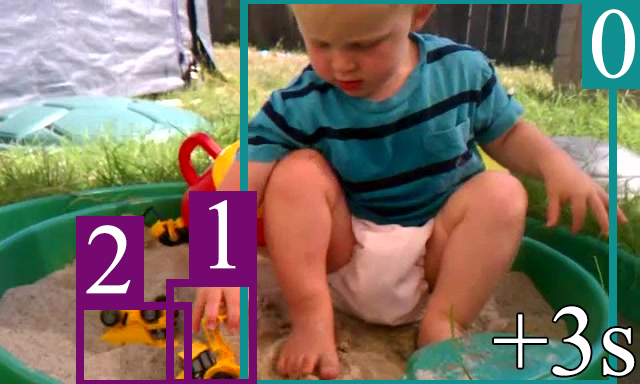}};
            \node[inner sep=0pt, right=2pt of f_3] (f_5) {\includegraphics[width=80pt]{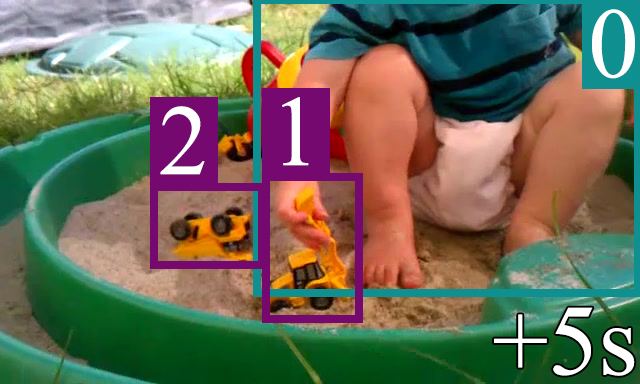}};
    
            \node[draw=none, rectangle, align=center, below left=-1pt and -30pt of f__1] (a1) {Anticipation $\tau_a=1$:};
            \node[draw=none, inner sep=0pt, rectangle, align=center, below=0pt of a1] (a11) {$\langle$\textcolor{darkcyan}{human0}, \textcolor{darkgreen}{\emph{next to}} \& \textcolor{darkgreen}{\emph{hold}} \& \textcolor{darkgreen}{\emph{watch}}, \textcolor{darkpurple}{toy1}$\rangle$};
            \node[draw=none, inner sep=0pt, rectangle, align=center, below=1pt of a11] (a12) {$\langle$\textcolor{darkcyan}{human0}, \textcolor{darkgreen}{\emph{next to}}, \textcolor{darkpurple}{toy2}$\rangle$};
    
            \node[draw=none, rectangle, align=center, below left=-1pt and -40pt of f_1] (a3) {Anticipation $\tau_a=3$:};
            \node[draw=none, inner sep=0pt, rectangle, align=center, below=0pt of a3] (a31) {$\langle$\textcolor{darkcyan}{human0}, \textcolor{darkgreen}{\emph{next to}} \& \textcolor{darkgreen}{\emph{hold}} \& \textcolor{darkgreen}{\emph{watch}}, \textcolor{darkpurple}{toy1}$\rangle$};
            \node[draw=none, inner sep=0pt, rectangle, align=center, below=1pt of a31] (a32) {$\langle$\textcolor{darkcyan}{human0}, \textcolor{darkgreen}{\emph{next to}}, \textcolor{darkpurple}{toy2}$\rangle$};
    
            \node[draw=none, rectangle, align=center, below left=-1pt and -50pt of f_5] (a5) {Anticipation $\tau_a=5$:};
            \node[draw=none, inner sep=0pt, rectangle, align=center, below=0pt of a5] (a51) {$\langle$\textcolor{darkcyan}{human0}, \textcolor{darkgreen}{\emph{next to}} \& \textcolor{darkgreen}{\emph{hold}} \& \textcolor{red}{\emph{watch}}, \textcolor{darkpurple}{toy1}$\rangle$};
            \node[draw=none, inner sep=0pt, rectangle, align=center, below=1pt of a51] (a52) {$\langle$\textcolor{darkcyan}{human0}, \textcolor{darkgreen}{\emph{next to}}, \textcolor{darkpurple}{toy2}$\rangle$};
            
        \end{tikzpicture}
        \caption{Qualitative results in \emph{Oracle} mode.}
        \label{fig:qualitative_oracle}
    \end{subfigure}
    
    \begin{subfigure}[]{0.999\textwidth}
        \centering
        \begin{tikzpicture}[font=\small, >={latex}]
            \node[draw=none, rectangle, align=center, minimum height=15pt, minimum width=80pt] (placeholder) {};
            \node[below=0pt of placeholder, inner sep=0pt] (f__2) {\includegraphics[width=80pt]{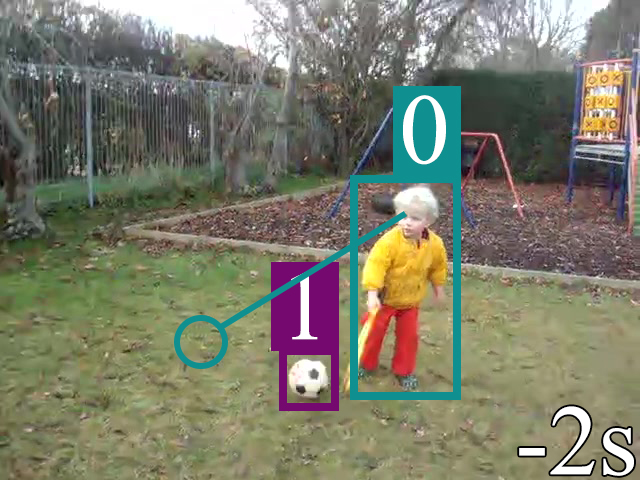}};
            \node[inner sep=0pt, right=2pt of f__2] (f__1) {\includegraphics[width=80pt]{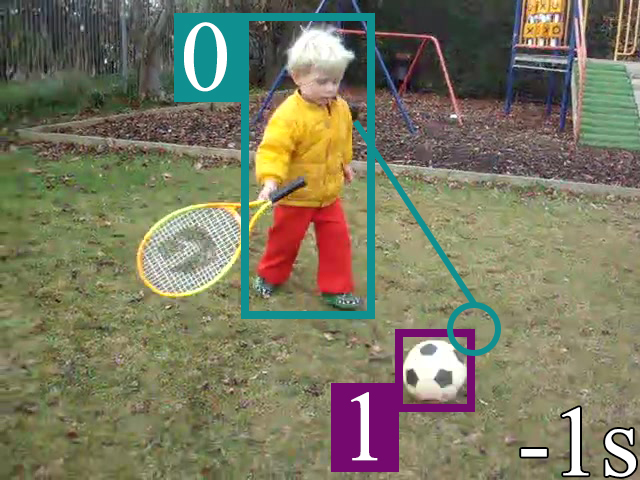}};
            \node[inner sep=0pt, right=2pt of f__1] (f_0) {\includegraphics[width=80pt]{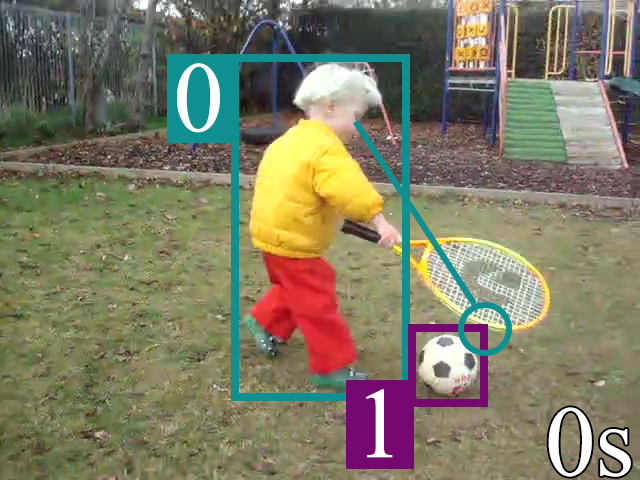}};
            \node[inner sep=0pt, right=2pt of f_0] (f_1) {\includegraphics[width=80pt]{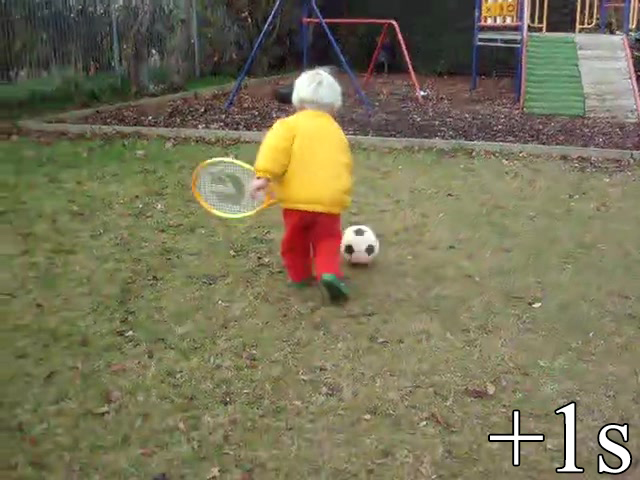}};
            \node[inner sep=0pt, right=2pt of f_1] (f_3) {\includegraphics[width=80pt]{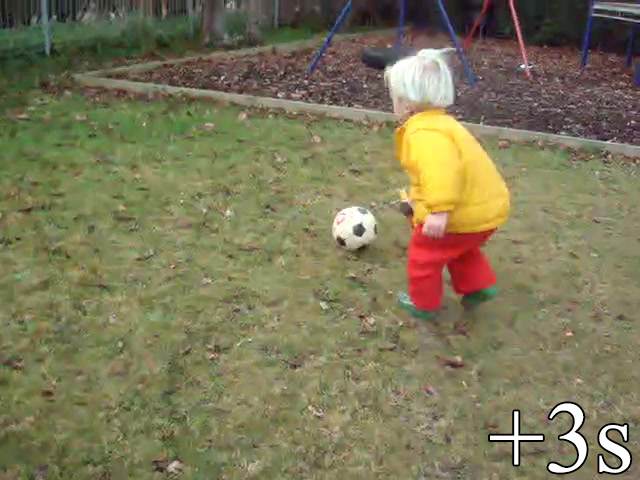}};
            \node[inner sep=0pt, right=2pt of f_3] (f_5) {\includegraphics[width=80pt]{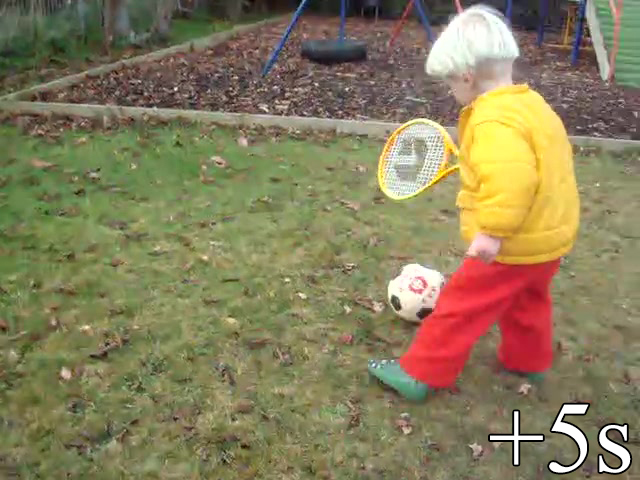}};
    
            \node[draw=none, rectangle, align=center, below left=-1pt and -35pt of f__1] (a1) {Anticipation $\tau_a=1$:};
            \node[draw=none, inner sep=0pt, rectangle, align=center, below=0pt of a1] (a11) {$\langle$\textcolor{darkcyan}{human0}, \textcolor{darkgreen}{\emph{next to}} \& \textcolor{darkgreen}{\emph{watch}}, \textcolor{darkpurple}{ball1}$\rangle$};
            \node[draw=none, inner sep=0pt, rectangle, align=center, below=1pt of a11] (a12) {$\langle$\textcolor{darkcyan}{human0}, \textcolor{darkgreen}{\emph{towards}}, \textcolor{darkpurple}{ball1}$\rangle$};
            \node[draw=none, inner sep=0pt, rectangle, align=center, below=1pt of a12] (a13) {$\langle$\textcolor{darkcyan}{human0}, \textcolor{gray}{\emph{next to}} \& \textcolor{gray}{\emph{hold}}, \textcolor{gray}{racket}$\rangle$};
    
            \node[draw=none, rectangle, align=center, below left=-1pt and -40pt of f_1] (a3) {Anticipation $\tau_a=3$:};
            \node[draw=none, inner sep=0pt, rectangle, align=center, below=0pt of a3] (a31) {$\langle$\textcolor{darkcyan}{human0}, \textcolor{darkgreen}{\emph{next to}} \& \textcolor{darkgreen}{\emph{watch}}, \textcolor{darkpurple}{ball1}$\rangle$};
            \node[draw=none, inner sep=0pt, rectangle, align=center, below=1pt of a31] (a32) {$\langle$\textcolor{darkcyan}{human0}, \textcolor{darkgreen}{\emph{towards}} \& \textcolor{red}{\emph{kick}}, \textcolor{darkpurple}{ball1}$\rangle$};
            \node[draw=none, inner sep=0pt, rectangle, align=center, below=1pt of a32] (a33) {};
    
            \node[draw=none, rectangle, align=center, below left=-1pt and -45pt of f_5] (a5) {Anticipation $\tau_a=5$:};
            \node[draw=none, inner sep=0pt, rectangle, align=center, below=0pt of a5] (a51) {$\langle$\textcolor{darkcyan}{human0}, \textcolor{darkgreen}{\emph{next to}} \& \textcolor{darkgreen}{\emph{watch}}, \textcolor{darkpurple}{ball1}$\rangle$};
            \node[draw=none, inner sep=0pt, rectangle, align=center, below=1pt of a51] (a52) {$\langle$\textcolor{darkcyan}{human0}, \textcolor{darkgreen}{\emph{towards}} \&  \textcolor{red}{\emph{kick}}, \textcolor{darkpurple}{ball1}$\rangle$};
            \node[draw=none, inner sep=0pt, rectangle, align=center, below=1pt of a52] (a53) {$\langle$\textcolor{darkcyan}{human0}, \textcolor{gray}{\emph{next to}} \& \textcolor{gray}{\emph{hold}}, \textcolor{gray}{racket}$\rangle$};
    
            \node[draw=none, rectangle, align=center, minimum height=3pt, minimum width=1pt, below=0pt of a13] (placeholder) {};
    
        \end{tikzpicture}
    
        \begin{tikzpicture}[font=\small, >={latex}]
    
            \node[inner sep=0pt] (f__2) {\includegraphics[width=80pt]{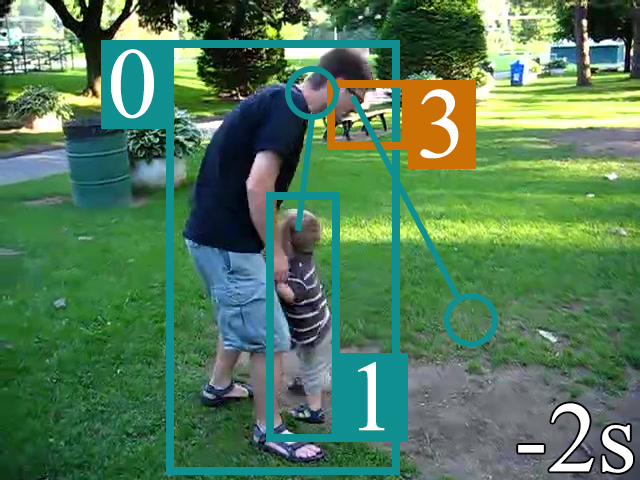}};
            \node[inner sep=0pt, right=2pt of f__2] (f__1) {\includegraphics[width=80pt]{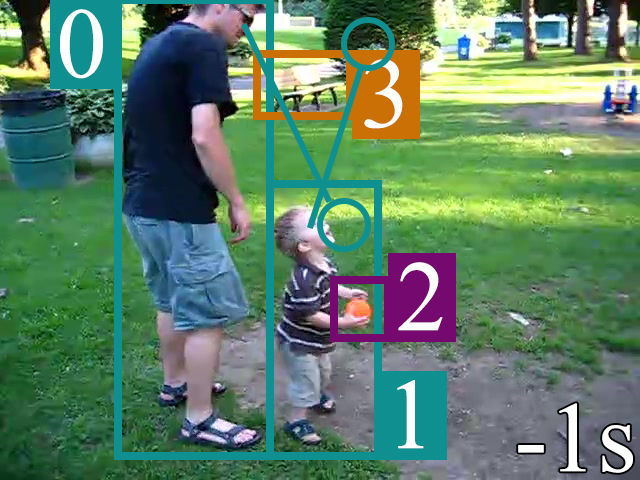}};
            \node[inner sep=0pt, right=2pt of f__1] (f_0) {\includegraphics[width=80pt]{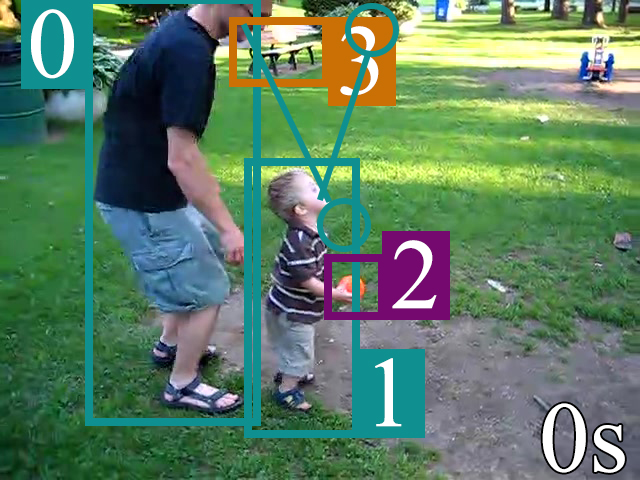}};
            \node[inner sep=0pt, right=2pt of f_0] (f_1) {\includegraphics[width=80pt]{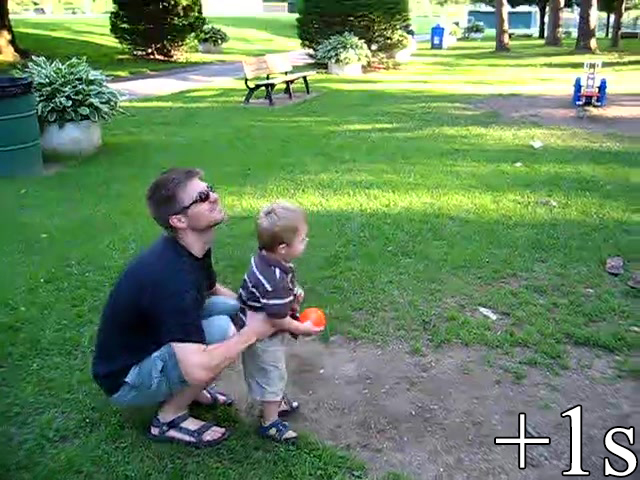}};
            \node[inner sep=0pt, right=2pt of f_1] (f_3) {\includegraphics[width=80pt]{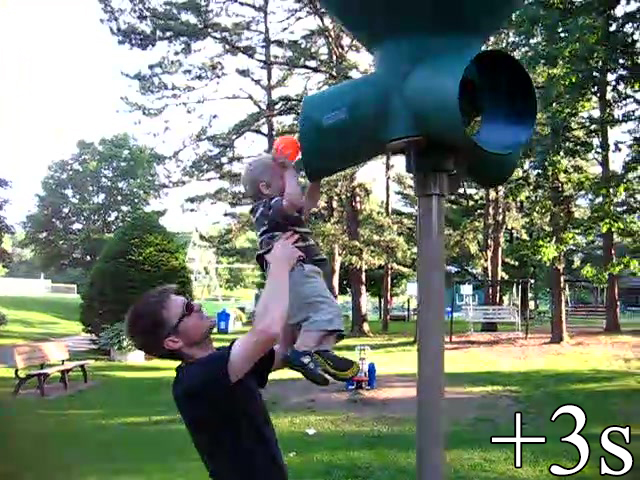}};
            \node[inner sep=0pt, right=2pt of f_3] (f_5) {\includegraphics[width=80pt]{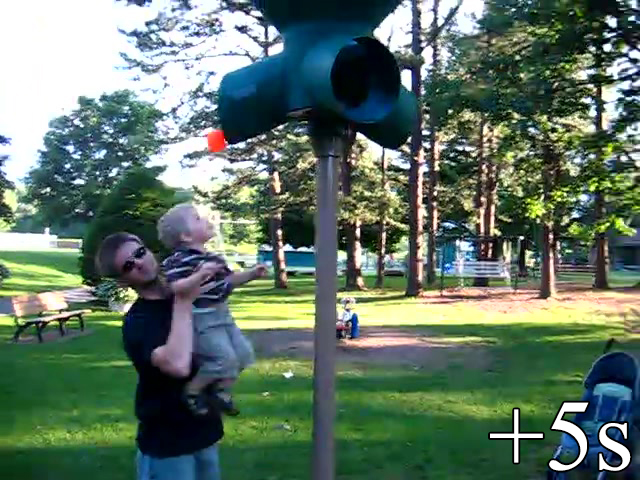}};
    
            \node[draw=none, rectangle, align=center, below left=-1pt and -35pt of f__1] (a1) {Anticipation $\tau_a=1$:};
            \node[draw=none, inner sep=0pt, rectangle, align=center, below=0pt of a1] (a11) {$\langle$\textcolor{darkcyan}{human0}, \textcolor{red}{\emph{in front of}} \& \textcolor{red}{\emph{watch}}, \textcolor{darkcyan}{human1}$\rangle$};
            \node[draw=none, inner sep=0pt, rectangle, align=center, below=1pt of a11] (a12) {$\langle$\textcolor{darkcyan}{human0}, \textcolor{gray}{\emph{behind}} \& \textcolor{gray}{\emph{hold}} \& \textcolor{gray}{\emph{lift}}, \textcolor{darkcyan}{human1}$\rangle$};
            \node[draw=none, inner sep=0pt, rectangle, align=center, below=1pt of a12] (a13) {$\langle$\textcolor{darkcyan}{human0}, \textcolor{darkgreen}{\emph{next to}}, \textcolor{darkpurple}{ball2}$\rangle$};
            \node[draw=none, inner sep=0pt, rectangle, align=center, below=1pt of a13] (a14) {$\langle$\textcolor{darkcyan}{human1}, \textcolor{darkgreen}{\emph{in front of}}, \textcolor{darkcyan}{human0}$\rangle$};
            \node[draw=none, inner sep=0pt, rectangle, align=center, below=1pt of a14] (a15) {$\langle$\textcolor{darkcyan}{human1}, \textcolor{darkgreen}{\emph{next to}}, \& \textcolor{darkgreen}{\emph{hold}}, \textcolor{darkpurple}{ball2}$\rangle$};
            \node[draw=none, inner sep=0pt, rectangle, align=center, below=1pt of a15] (a16) {$\langle$\textcolor{darkcyan}{human1}, \textcolor{red}{\emph{towards}}, \textcolor{darkorange}{bench3}$\rangle$};

            \node[draw=none, rectangle, align=center, below left=-1pt and -40pt of f_1] (a3) {Anticipation $\tau_a=3$:};
            \node[draw=none, inner sep=0pt, rectangle, align=center, below=0pt of a3] (a31) {$\langle$\textcolor{darkcyan}{human0}, \textcolor{red}{\emph{next to}} \& \textcolor{red}{\emph{watch}}, \textcolor{darkcyan}{human1}$\rangle$};
            \node[draw=none, inner sep=0pt, rectangle, align=center, below=1pt of a31] (a32) {$\langle$\textcolor{darkcyan}{human0}, \textcolor{darkgreen}{\emph{behind}} \& \textcolor{gray}{\emph{hold}} \& \textcolor{gray}{\emph{lift}}, \textcolor{darkcyan}{human1}$\rangle$};
            \node[draw=none, inner sep=0pt, rectangle, align=center, below=1pt of a32] (a33) {$\langle$\textcolor{darkcyan}{human0}, \textcolor{darkgreen}{\emph{next to}}, \textcolor{darkpurple}{ball2}$\rangle$};
            \node[draw=none, inner sep=0pt, rectangle, align=center, below=1pt of a33] (a34) {$\langle$\textcolor{darkcyan}{human1}, \textcolor{darkgreen}{\emph{in front of}}, \textcolor{darkcyan}{human0}$\rangle$};
            \node[draw=none, inner sep=0pt, rectangle, align=center, below=1pt of a34] (a35) {$\langle$\textcolor{darkcyan}{human1}, \textcolor{darkgreen}{\emph{next to}}, \& \textcolor{darkgreen}{\emph{hold}} \& \textcolor{gray}{\emph{lift}}, \textcolor{darkpurple}{ball2}$\rangle$};
            \node[draw=none, inner sep=0pt, rectangle, align=center, below=1pt of a35] (a36) {$\langle$\textcolor{darkcyan}{human1}, \textcolor{red}{\emph{towards}}, \textcolor{darkorange}{bench3}$\rangle$};
    
            \node[draw=none, rectangle, align=center, below left=-1pt and -45pt of f_5] (a5) {Anticipation $\tau_a=5$:};
            \node[draw=none, inner sep=0pt, rectangle, align=center, below=0pt of a5] (a51) {$\langle$\textcolor{darkcyan}{human0}, \textcolor{red}{\emph{next to}} \& \textcolor{red}{\emph{watch}}, \textcolor{darkcyan}{human1}$\rangle$};
            \node[draw=none, inner sep=0pt, rectangle, align=center, below=1pt of a51] (a52) {$\langle$\textcolor{darkcyan}{human0}, \textcolor{darkgreen}{\emph{behind}} \& \textcolor{gray}{\emph{hold}}, \textcolor{darkcyan}{human1}$\rangle$};
            \node[draw=none, inner sep=0pt, rectangle, align=center, below=1pt of a52] (a53) {$\langle$\textcolor{darkcyan}{human0}, \textcolor{red}{\emph{next to}}, \textcolor{darkpurple}{ball2}$\rangle$};
            \node[draw=none, inner sep=0pt, rectangle, align=center, below=1pt of a53] (a54) {$\langle$\textcolor{darkcyan}{human1}, \textcolor{darkgreen}{\emph{in front of}} \& \textcolor{red}{\emph{away}}, \textcolor{darkcyan}{human0}$\rangle$};
            \node[draw=none, inner sep=0pt, rectangle, align=center, below=1pt of a54] (a55) {$\langle$\textcolor{darkcyan}{human1}, \textcolor{red}{\emph{next to}}, \& \textcolor{red}{\emph{hold}} \& \textcolor{darkgreen}{\emph{watch}}, \textcolor{darkpurple}{ball2}$\rangle$};
            \node[draw=none, inner sep=0pt, rectangle, align=center, below=1pt of a55] (a56) {$\langle$\textcolor{darkcyan}{human1}, \textcolor{red}{\emph{away}}, \textcolor{darkorange}{bench3}$\rangle$};
    
        \end{tikzpicture}
        \caption{Qualitative results in \emph{Detection} mode. }
            \label{fig:qualitative_detection}
    \end{subfigure}

    \caption{Qualitative results of \textbf{HOI anticipation} task on the VidHOI dataset~\citep{hoi_v_set:VidHOI}. Our model observes six past frames (three are shown in the figure) and produces HOI anticipations for anticipation time gap $\tau_a=\{1, 3, 5\}$ seconds. \textcolor{darkgreen}{Green}: true positives. \textcolor{red}{Red}: false positives. \textcolor{gray}{Gray}: false negatives.}
    \label{fig:qualitative}
\end{figure*}

To further investigate the performance of our model, we show the qualitative results for the HOI anticipation task in $Oracle$ mode in Figure~\ref{fig:qualitative_oracle}. For simplification, we show only the top-$5$ results for one human in each scene. In the upper scene, our model forecasts that the \textcolor{darkcyan}{human0} will \emph{wave} the \textcolor{darkpurple}{bat1} at any time in the future, which is logical. In the bottom scene, the gaze cues can probably help the model to understand that the baby is focusing on the \textcolor{darkpurple}{toy1} and will not play with another toy in the near future. 

Figure~\ref{fig:qualitative_detection} shows two more HOI anticipation results on the VidHOI dataset~\citep{hoi_v_set:VidHOI} in \emph{Detection} mode. In the first scene, our model predicts that the child is going to \emph{kick} the \textcolor{darkpurple}{ball}. However, in fact, the child is playing the ball with a racket. Our object detector fails to recognize that racket, thus, our spatio-temporal transformer is unable to fully understand the scene. When we provide the model with the ground-truth object annotations, it does not produce the triplet $\langle$\textcolor{darkcyan}{human0}, \textcolor{black}{\emph{kick}}, \textcolor{darkpurple}{ball1}$\rangle$. In the second video clip, our framework successfully detects the necessary objects to understand the scene. Nevertheless, it still cannot forecast that the adult will \emph{lift} the child and the child will \emph{lift} the ball. This is also hard to predict for us humans since the interactions between two humans are more uncertain in the future. In addition, the bench detected in the background is irrelevant to the two humans. However, the gaze direction of the child estimated by the gaze-following model is roughly in the direction of the bench. The transformer may capture misleading contexts that could affect the model performance. Thus, overall, the gaze cue is a useful feature, but there is room to improve its usage. 

\subsection{Ablation Study}

\begin{table}
    \begin{subtable}[]{0.47\textwidth}
        \centering
        \small
        \begin{tabular}{|c|c|c|c c c|}
            \hline
            \multirow{2}{*}{Setting} & \multirow{2}{*}{$L$} & \multirow{2}{*}{Gaze} & \multicolumn{3}{c|}{mAP} \\
            \cline{4-6}
            & & & Full & Non-rare & Rare \\
            \hhline{|=|=|=|= = =|}
            STTran & 2 & Concat & 28.58 & 42.00 & 18.28 \\
            + MLM → CB & = & = & 34.20 & 46.92 & 24.43 \\
            + WS & = & = & 34.97 & 47.85 & 25.08 \\
            + HF & = & = & 35.22 & 48.01 & 25.40\\
            + PW & = & = & 35.39 & 48.82 & 25.07 \\
            = & 4 & = & 35.20 & 48.84 & 24.73 \\
            \textbf{=} & \textbf{6} & \textbf{=} & \textbf{36.29} & \textbf{49.43} & \textbf{26.19} \\
            = & 8 & = & 35.88 & 49.29 & 25.58 \\
            \hline
            = & 6 & Cross & 36.78 & 50.48 & 26.25 \\
            + PW → IW & = & = & 37.85 & 51.09 & 27.68 \\
            + G & = & = & 38.35 & 52.30 & 27.63 \\
            + Learned → Sine & = & = & 38.49 & 52.17 & 27.98 \\
            \underline{+ WD} & \underline{=} & \underline{=} & \underline{\textbf{38.61}} & \underline{\textbf{52.44}} & \underline{\textbf{27.99}} \\
            \hline
        \end{tabular}
        \caption{Ablation study for our improvements to STTran~\citep{hoi_v2:sttran}}
        \label{table:ablation_components}
    \end{subtable}

    \begin{subtable}[]{0.47\textwidth}
        \centering
        \small
        \begin{tabular}{|c|c|c|c c c c|}
            \hline
            \multirow{2}{*}{$\tau_a$} & \multirow{2}{*}{Gaze} & mAP & \multicolumn{4}{c|}{Person-wise top-5} \\
            \cline{3-7}
            & & Full & Rec & Prec & Acc & F1 \\
            \hhline{|=|=|=|= = = =|}
            \multirow{2}{*}{0} & w/o & 37.27 & 69.48 & \textbf{60.75} & \textbf{51.68} & 62.24 \\
            & Cross & \textbf{38.61} & \textbf{70.91} & 59.84 & 51.29 & \textbf{62.24} \\
            \hhline{|=|=|=|= = = =|}
            \multirow{2}{*}{1} & w/o & 36.14 & 70.92 & 59.93 & 51.37 & 62.28 \\
            & Cross & \textbf{37.59} & \textbf{72.17} & \textbf{59.98} & \textbf{51.65} & \textbf{62.78}\\
            \hline
            \multirow{2}{*}{3} & w/o & 32.55 & 70.37 & 59.67 & 51.09 & 61.90 \\
            & Cross & \textbf{33.14} & \textbf{71.88} & \textbf{60.44} & \textbf{52.08} & \textbf{62.87}\\
            \hline
            \multirow{2}{*}{5} & w/o & 32.05 & 69.03 & \textbf{59.38} & 50.72 & 61.24 \\
            & Cross & \textbf{32.75} & \textbf{71.25} & 59.09 & \textbf{51.14} & \textbf{61.92}\\
            \hline
            \multirow{2}{*}{7} & w/o & 31.32 & 69.18 & \textbf{59.50} & \textbf{50.67} & 61.24 \\
            & Cross & \textbf{31.70} & \textbf{70.48} & 58.80 & 50.56 & \textbf{61.36}\\
            \hline
        \end{tabular}
        \caption{Ablation study for our model with or without (w/o) gaze}
        \label{table:ablation_gaze}
    \end{subtable}
    
    \caption{Ablation study on our framework. All experiments are conducted in \emph{Oracle} mode. The underlined setting is adopted for comparison with the baselines. ``='' represents the same option as in the previous row. MLM: multi-label margin loss. CB: class-balanced focal loss. WS: sampling batch of windows from different videos. HF: random horizontal flipping as data augmentation. PW: person-wise sliding window in the temporal encoder. IW: pair-wise (instance-wise) sliding window. G: using the global token in the spatial encoder. Learned: learned positional encoding. Sine: sinusoidal positional encoding. WD: applying weight decay in AdamW optimizer. L: length of the sliding window. Concat: Concatenating human gaze features with pair representations, no cross-attention in the temporal encoder. Cross: Using human gaze features as keys and values in the cross-attention mechanism.}
    \label{table:ablation}
\end{table}

We conduct an extensive ablation study to investigate the effectiveness of gaze features and our improvements to the STTran model~\citep{hoi_v2:sttran}. The experiments for different tricks and components are performed on the HOI detection task. The best setup is applied to anticipation tasks with all anticipation times. We first examine the usage of gaze cues as an additional component in the human-object relation representations, i.e., we concatenate the gaze feature with the visual appearance, spatial relation, and semantic feature in the input embedding block. The temporal encoder only contains stacked self-attention layers. The dependencies between the human gaze and other features are then extracted solely through the self-attention mechanism. We apply the setting with the highest mAP Full as the base setting for further experiments using gaze features in cross-attention layers.

Table~\ref{table:ablation_components} shows that all of our modified or added components are able to increase the mAP Full. Changing the loss function from multi-label margin (MLM) loss to CB Focal loss improves our model performance the most. The rare mAP is increased by $33.6\%$. This observation meets our aim of applying CB Focal loss, which should address the challenge of extreme dataset imbalance. By increasing window length to $6$, the model achieves overall the best performance in gaze concatenation mode. However, further raising the window length to $8$ instead reduces the mAP. This performance drop might be caused by the fact that a longer window of frames may capture more temporal information which is no more related to the current interactions. After changing the gaze usage from concatenation to cross-attention, our model gains further performance boost. More experiments also confirm that the pair-wise sliding window and the explicit global context are beneficial for HOI detection from videos.

Finally, in Table~\ref{table:ablation_gaze}, we show that the gaze features are beneficial for both HOI detection and anticipation tasks. However, the performance improvement is not as significant as we expect. The main reason could be that the spatio-temporal transformer is trained with noisy gaze cues as the VidHOI dataset lacks ground-truth gaze annotations. The performance of the adopted gaze following model~\citep{gaze:detecting_attended} might be a limitation of our framework, but could be improved by leveraging more recent works in that field, such as~\citep{gaze:end_to_end, gaze:dam}. In addition, even though the gaze does not result in big improvement, other extensions we proposed in the spatio-temporal transformer still boost the model performance and allow us to achieve state-of-the-art in HOI detection and anticipation in videos.

\section{Conclusion}
In this work, we propose a multimodal framework to detect and anticipate HOIs from a third-person video by additionally leveraging gaze cues in the cross-attention mechanism. We utilize an object tracker to enable the temporal encoder to focus on the temporal evolution of each human-object pair separately. Addressing the extreme dataset imbalance issue in VidHOI dataset~\citep{hoi_v_set:VidHOI}, we adopt the class-balanced Focal loss. Furthermore, we propose a person-wise multi-label criterion to evaluate the models in HOI anticipation tasks in multi-person scenarios. Experimental results demonstrate that our framework outperforms the current state-of-the-art for HOI detection and anticipation tasks on the VidHOI dataset and the gaze features are beneficial to both tasks. For future works, adding more modalities such as depth information or human pose features could be advantageous. Furthermore, based on the HOI anticipation results, policies could be developed for human-assistive robots. 

\section*{Acknowledgments}
This work is funded by Marie Sklodowska-Curie Action Horizon 2020 (Grant agreement No. 955778) for project ``Personalized Robotics as Service Oriented Applications'' (PERSEO).

\bibliographystyle{model2-names}
\bibliography{refs}

\end{document}